\documentclass[runningheads]{llncs}
\usepackage{graphicx}

\usepackage{tikz}
\usepackage{comment}
\usepackage{amsmath,amssymb} 
\usepackage{color}

\usepackage{amsmath}
\usepackage{amssymb}
\usepackage{booktabs}
\usepackage{makecell}
\usepackage{multicol,multirow}
\usepackage{microtype}
\usepackage{enumitem}
\usepackage{algorithm}
\usepackage{algorithmic}
\usepackage{alltt}
\usepackage{wrapfig}
\usepackage{hyperref}

\usepackage[accsupp]{axessibility}  

\def\0{{\bf 0}}
\def\1{{\bf 1}}

\def\ie{\text{i.e.}}
\def\eg{\text{e.g.}}

\def\eg{\emph{e.g.}} 

\def\ie{\emph{i.e.}}

\def\etc{\emph{etc.}}

\definecolor{OliveGreen}{RGB}{34, 139, 34}
\definecolor{Black}{RGB}{0, 0, 0}

\begin{document}
\pagestyle{headings}
\mainmatter
\def\ECCVSubNumber{3302}  

\title{Contrastive Vision-Language Pre-training with Limited Resources} 


\titlerunning{Contrastive Vision-Language Pre-training with Limited Resources}
%
\author{Quan Cui\inst{1,2} \and
Boyan Zhou\inst{1} \and
Yu Guo\inst{1} \and 
Weidong Yin\inst{1} \and \\
Hao Wu\inst{1}\thanks{Corresponding author.} \and
Osamu Yoshie\inst{2} \and
Yubo Chen\inst{1}
}
\authorrunning{Q. Cui et al.}
%
\institute{ByteDance \and Waseda University \\
\email{cui-quan@toki.waseda.jp, wuhao.5688@bytedance.com}}
\maketitle

\begin{abstract}
Pioneering dual-encoder pre-training works~(\eg, CLIP and ALIGN) have revealed the potential of aligning multi-modal representations with contrastive learning. However, these works require a tremendous amount of data and computational resources~(\eg, billion-level web data and hundreds of GPUs), which prevent researchers with limited resources from reproduction and further exploration. To this end, we propose a stack of novel methods, which significantly cut down the heavy resource dependency and allow us to conduct dual-encoder multi-modal representation alignment with limited resources. Besides, we provide a reproducible baseline of competitive results, namely ZeroVL, with only 14M publicly accessible academic datasets and 8 V100 GPUs. Additionally, we collect 100M web data for pre-training, and achieve comparable or superior results than state-of-the-art methods, further proving the effectiveness of our methods on large-scale data. We hope that this work will provide useful data points and experience for future research in contrastive vision-language pre-training. Code is available at \href{https://github.com/zerovl/ZeroVL}{https://github.com/zerovl/ZeroVL}. 
\end{abstract}

\section{Introduction}

Large-scale representation pre-training has become the de-facto approach in vision~\cite{simclr,simclrv2,moco,wsl}, language~\cite{bert,roberta,deberta} and vision-language~\cite{clip,align} modeling tasks. In the vision-language pre-training field, most mainstream approaches fall into one of two classes: single-encoder~\cite{videobert,visualbert,vl-bert,uniter,oscar,vinvl,villa,soho,vilt,abf} and dual-encoder~\cite{clip,align}. Typical single-encoder approaches focus on learning semantic alignments between image regions and text entities with a single backbone network, greatly benefiting various downstream multi-modal tasks, \eg, VQA~\cite{vqa1,vqa2,vqa3}, VCR~\cite{vcr} and NLVR~\cite{nlvr1,nlvr2}, \etc.
In real-scenario applications~\cite{styleclip}, dual-encoder pre-training approach could be preferable for its flexibility. For one thing, downstream tasks of either modality can benefit from the pre-training. For another, dual-encoder approaches are more efficient than single-encoder approaches on popular multi-modal industrial applications, \eg, cross-modal matching and retrieval tasks~\cite{gpo,pq}.

Recent works~\cite{clip,align} have demonstrated that, by aligning visual and language representations with the contrastive loss, a simple dual-encoder architecture is able to yield state-of-the-art representation learning performances.
However, we notice a significant problem which might obstruct the progress in this research direction, \ie, pioneering works require a tremendous amount of vision-linguistic corpus and computational resources for training, and such heavy resource dependency prevents researchers with limited resources from reproduction and further explorations. For instance, CLIP~\cite{clip} and ALIGN~\cite{align} respectively collected 400M and 1.8B web image-text pairs and trained models with 256 V100 GPUs and 1,024 TPU cores. Such experimental environments present a big challenge for the most researchers, and further lead to a lack of commonly reproducible benchmarks for dual-encoder model, making it hard to validate novel methods.

\begin{table}[t]
\centering
\setlength{\tabcolsep}{0.86mm}{
\begin{scriptsize}
\begin{tabular}{l|cc|c|cc|cc} \Xhline{2\arrayrulewidth}
\multirow{2}{*}{method} &  \multicolumn{2}{c|}{\textbf{computation}} & \multirow{2}{*}{\textbf{data}} & \multicolumn{2}{c|}{MS-COCO}    & \multicolumn{2}{c}{F30K}      \\
                        &     device        & count &             & zs. & ft. & zs. & ft. \\ \hline
CLIP~\cite{clip}           &  V100 &  256       & 400M                    & 400.2 &  --  & 540.6          &  --                 \\
ALIGN~\cite{align}         & TPU$_{\text{v3}}$ &  1,024        & 1800M                  & 425.3 &  500.4   & \textbf{553.3}          &   \textbf{576.0}                 \\ \hline
baseline           & V100 &  8   & 14.2M                  &  371.6 &  471.9    & 483.3          &   553.0                 \\
ZeroVL       & V100 & 8 & 14.2M                  &  425.0 & 485.0    & 536.2          &    561.6                 \\
ZeroVL$\dagger$       & V100 & 8 & 100M            & \textbf{442.1}  &  \textbf{500.5}         & 546.5          &     573.6           \\ \Xhline{2\arrayrulewidth}   
\end{tabular}
\end{scriptsize}}
\caption{ \footnotesize{Statistics of training resources and cross-modal retrieval RSUM scores~\cite{gpo,rsum1,rsum2}. ``zs.'' and ``ft.'' represent zero-shot and fine-tuned settings. ``$\dagger$'' means pre-training with 100M web data.}}
\label{tab:introduction}
\end{table}

To alleviate the problems above, we design a comprehensive training pipeline with only open-source academic datasets and limited computational resources. Specifically, we propose a collection of novel methods to deal with limited data and computation, respectively. Our proposed methods boost model performances while only introducing marginal overhead to both computation and implementation. As shown in Table~\ref{tab:introduction}, we achieve competitive results with $\sim$14M academic data and 8 V100 GPUs, greatly alleviating the heavy dependency on data and computation of contrastive language-image pre-training. 
To further demonstrate the effectiveness of our method on large-scale data, we collect 100M web image-text images and conduct pre-training without fine-tuning hyper-parameters. Surprisingly, our method successfully outperforms CLIP and achieves comparable results with ALIGN on pre-training and fine-tuning tasks.

\section{A Naive Baseline}
\label{sec:dataset}
In this section, we build up a naive baseline for stacking our methods and polishing it to a strong one. Methods are related to \textit{training with limited data} and \textit{training with limited computation resource}, which will be discussed in Sec.~\ref{sec:data} and~\ref{sec:training}.

\subsection{Pre-Training Datasets}
To ensure reproducibility, only publicly accessible academic datasets are leveraged to demonstrate the effectiveness of our methods. The statistics of collected image-text pair datasets are reported in Table~\ref{tab:statistics}. Four widely-used image-text pair datasets are selected for pre-training, \ie, \textit{(1)~SBU Captioned Photos~(SBU)}~\cite{sbu}, \textit{(2)~Visual Genome~(VG)}~\cite{vg}, \textit{(3)~Conceptual Captions 3M~(CC3M)}~\cite{cc3m}, and \textit{(4)~Conceptual 12M~(CC12M)}~\cite{cc12m} datasets. Detailed introductions are attached in the appendix.

\begin{table}[t] 
\centering
\setlength{\tabcolsep}{0.6mm}{
\begin{scriptsize}
\begin{tabular}{l|ccccc|cc} \Xhline{2\arrayrulewidth}
         & \multicolumn{5}{c|}{Pre-training} & \multicolumn{2}{c}{Test} \\ 
         & Total & SBU      & VG      & CC3M     & CC12M     & MS-COCO      & F30K            \\ \hline
\#image & 14.23M & 0.86M     & 0.50M    & 2.81M    & 10.06M    & 5K              & 1K              \\
\#text  & 14.23M & 0.86M     & 0.50M    & 2.81M    & 10.06M    & 25K             & 5K              \\ \Xhline{2\arrayrulewidth}
\end{tabular}
\end{scriptsize}}
\caption{The statistics of datasets for pre-training and test.}
\label{tab:statistics}
\end{table}

\subsection{Baseline Settings}
\label{subsec:baseline_setting}
Baseline settings are elaborated from the data, model, and training perspectives.
\begin{wrapfigure}{r}{0.4\linewidth}
	\centering
    \includegraphics[width=1.0\linewidth]{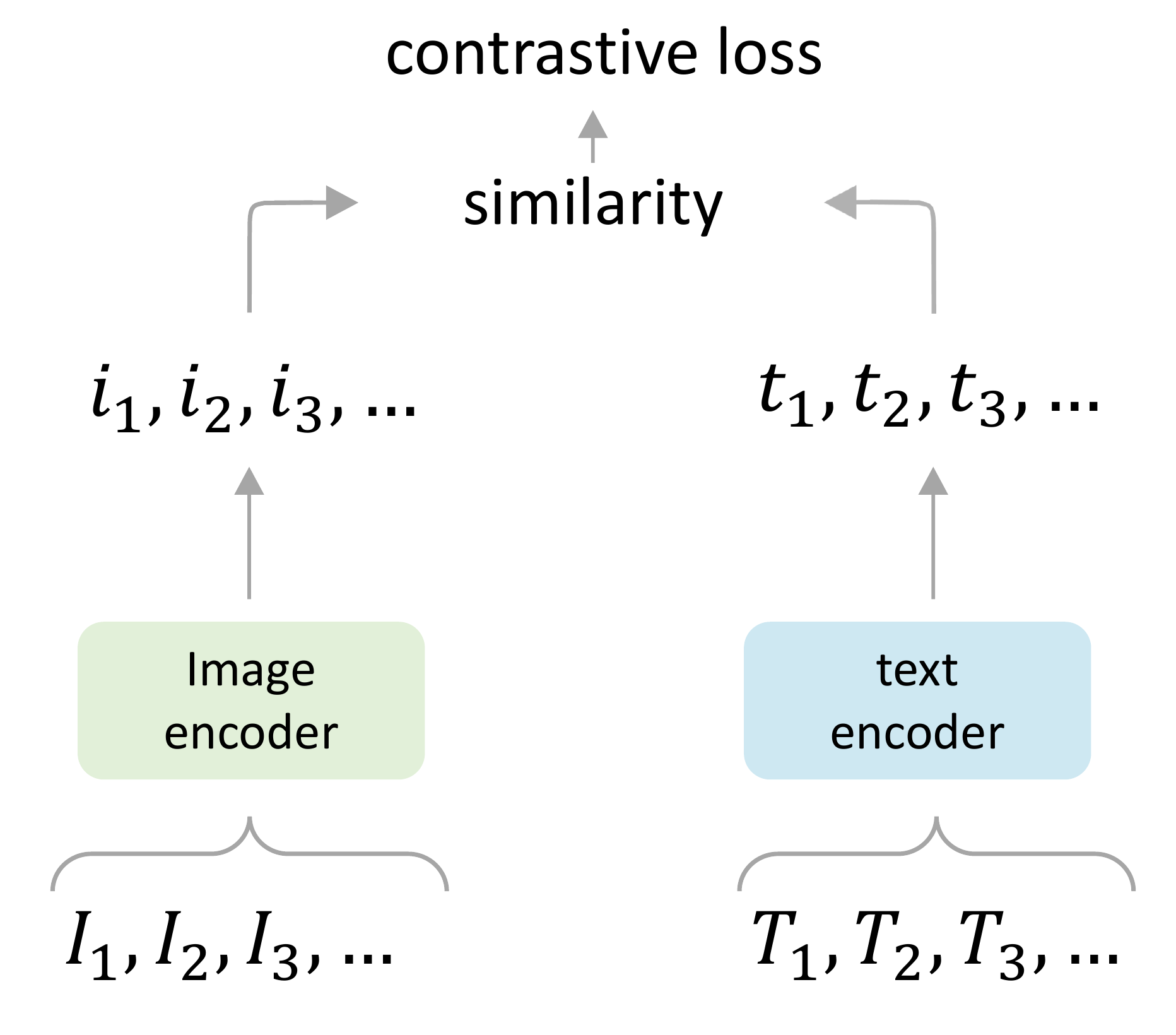}
	\caption{Illustration of the dual-encoder model architecture.} 
	\label{fig:framework}
\end{wrapfigure}

\noindent \textbf{Data preparation.} 
Batches are comprised by randomly sampling image-text pairs from pre-training datasets. 
Following~\cite{clip,align}, each image is randomly cropped to a rectangular region with aspect ratio sampled in $[3/4,4/3]$ and area sampled in $[60\%,100\%]$, then resized to 224$\times$224 resolution. 
Regarding the corresponding text, we use a percentage of 20\% input words for processing. For each word, we mask it, replace it with a random word, or delete it with a probability of 50\%, 10\% and 40\%, respectively.
During test, images are resized to 256$\times$256 and center cropped to 224$\times$224, while no specific process is applied to texts.

\noindent \textbf{Model architecture.} Inspired by~\cite{clip,align}, we employ a simple dual-encoder model to align visual and language representations of image-text pairs via a contrastive loss. The framework is illustrated in Figure~\ref{fig:framework}. Image and text encoders are ViT-B/16~\cite{vit} and BERT-Base~\cite{bert}, respectively. \texttt{[CLS]} tokens from image and text encoders are extracted and then projected to compact embeddings for calculating the contrastive loss.

\noindent \textbf{Training.}
AdamW~\cite{adam,adamw} optimizer is used for training and the weight decay is 1e-3. The dual-encoder model is trained for 20 epochs on 8 Nvidia V100 GPUs with a batch size of 1,024. The learning rate is initialized to 1e-4 and follows a cosine decay schedule. Notably, we set a minimum learning rate 1e-5 to avoid over-fitting. The embedding dimension for image and text representations is 512 and the trainable temperature of contrastive loss is initialized to 0.02.

\subsection{Evaluations}

\noindent \textbf{Metrics. }
Typically, multi-modal retrieval tasks are assessed with the recall at K (R@K) metric, with $\text{K}=\{1,5,10\}$. We follow~\cite{rsum1,rsum2,gpo} to use RSUM as the metric to reveal the overall performance, which is defined as the sum of recall metrics at $\text{K}=\{1,5,10\}$ of both image-to-text and text-to-image retrieval tasks.

\noindent \textbf{Test datasets. } Following the standard practice in~\cite{clip,align,gpo,vsepp,rsum1,rsum2}, we evaluate representations of pre-trained models by carrying out \textit{zero-shot} image-text retrieval tasks on test sets of \textit{(1)~MS-COCO Captions Karpathy's split~(MS-COCO)} and \textit{(2)~Flickr 30K~(F30K)} datasets. MS-COCO and F30K results are reported with 5K and 1K test images, respectively.


\section{Training with Limited Data Resource}
\label{sec:data}

Due to the copyright or technical issues, publicly accessible image-text academic datasets are greatly limited. The common practice to construct vision-linguistic corpus is collecting datasets from multiple sources. However, it brings in the dataset bias issue, which is caused by different collection manners of these datasets. Besides, limited data could suffer from the over-fitting problem, and seldom efforts were made for creating extra data for multi-modal pre-training.
In this section, we study how to take full advantages of limited data from these two perspectives, \ie, (1)~leveraging biased data and (2)~creating extra data. 

\subsection{Leveraging Biased Data with Debiased Sampling}

\begin{figure}[!htb]
    \centering
    \begin{minipage}{.55\textwidth}
        \centering
        \includegraphics[width=1.0\linewidth]{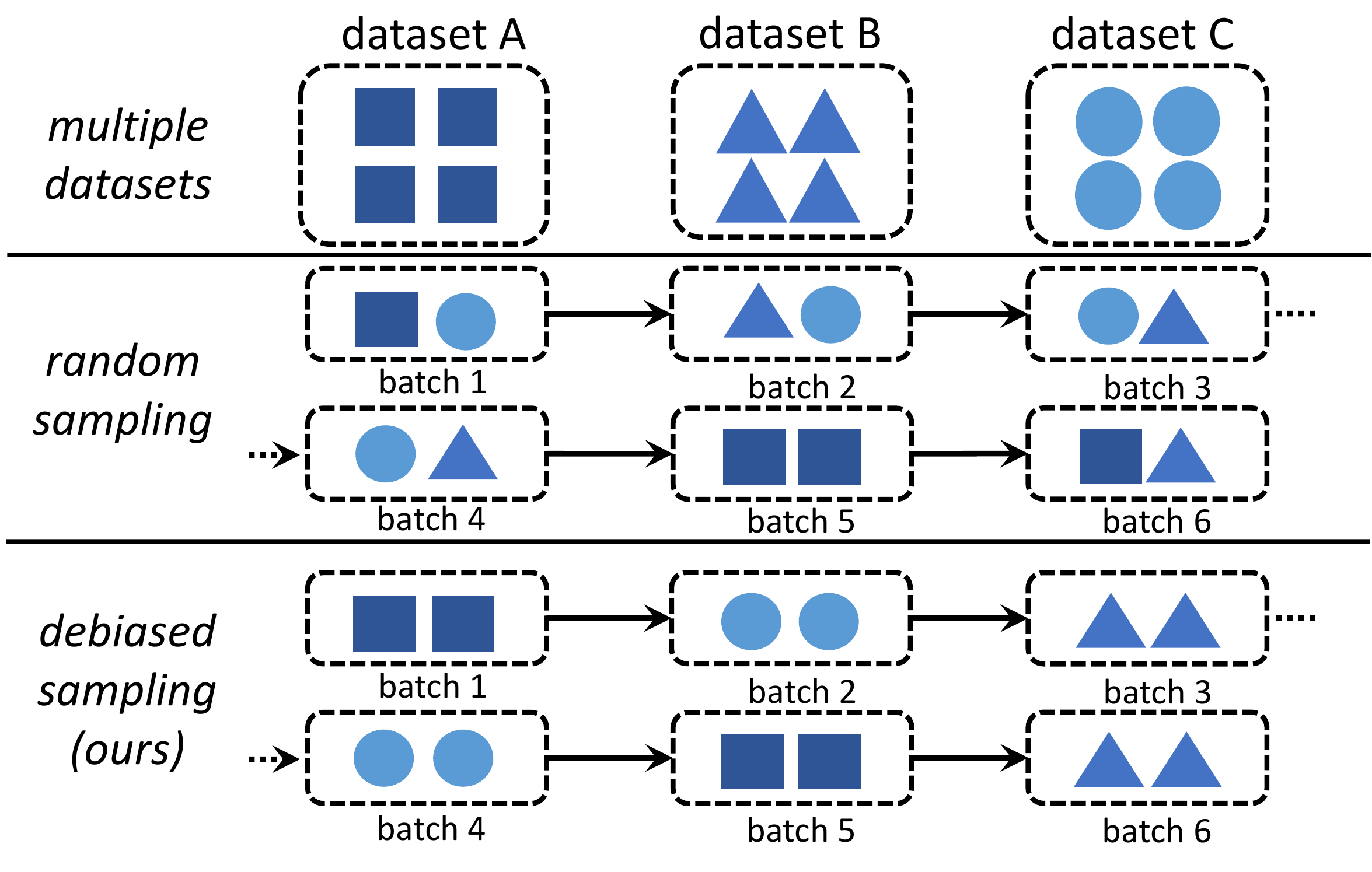}
        \caption{Illustration of sampling strategies.}
        \label{fig:sampling}
    \end{minipage}%
    \quad 
    \begin{minipage}{0.4\textwidth}
        \centering
        \includegraphics[width=1.0\linewidth]{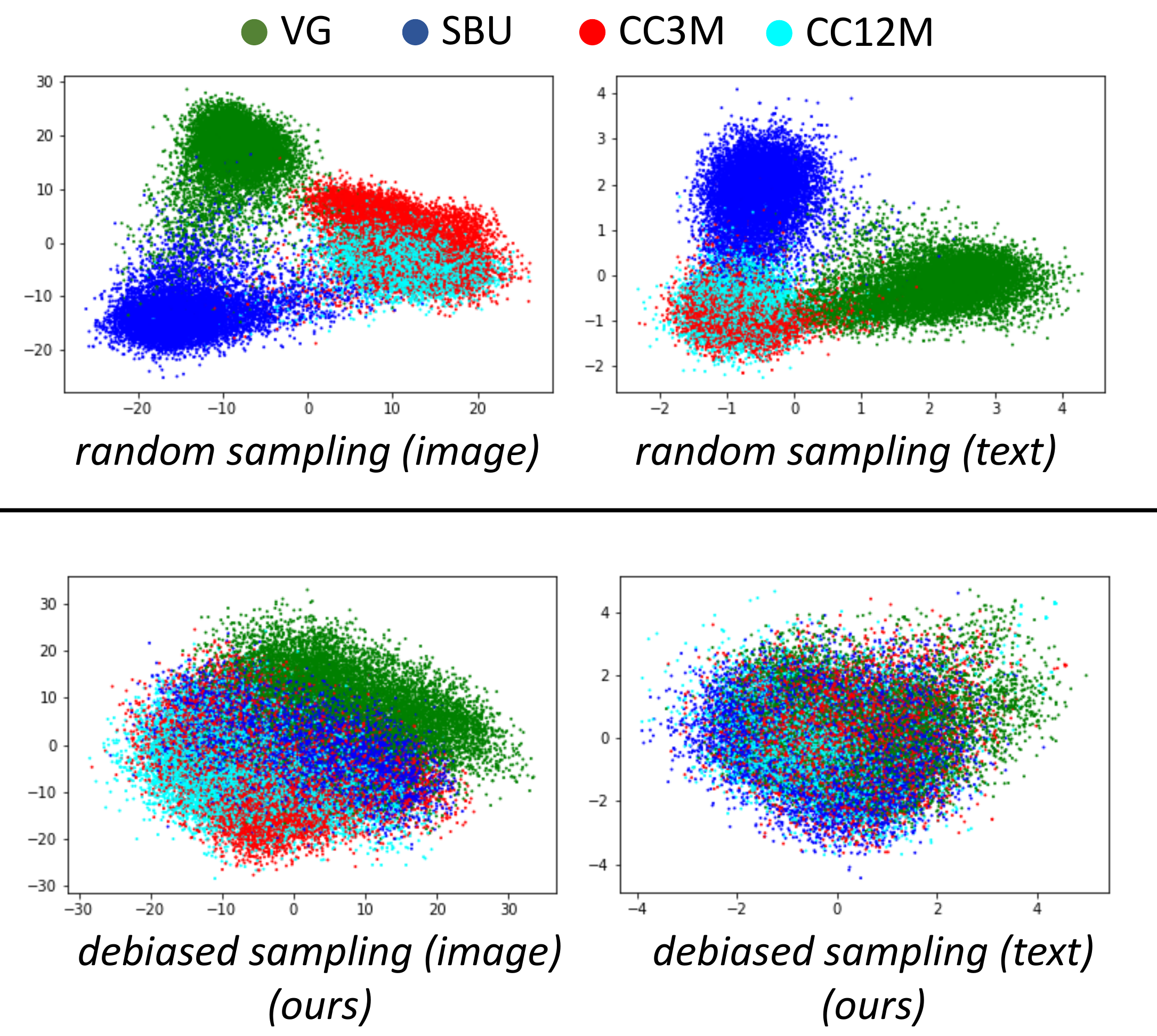}
        \caption{Illustration of image and text embeddings.}
        \label{fig:embeddings}
    \end{minipage}
\end{figure}

\noindent \textbf{Random sampling brings in dataset bias.}
Random sampling is an intuitive and widely used strategy, which randomly constructs training batches with all available data, as illustrated in Figure~\ref{fig:sampling}. However, when a batch is composed of samples from different datasets, models could be driven to distinguish negative samples by hacking the source information, \ie, learning the dataset bias. For instance, dataset A is mainly composed of \textit{natural scenery photos with long captions}, while dataset B is mainly comprised of \textit{people with short captions}. To distinguish samples from A and B, models are allowed to remember the dataset bias on image contents and caption lengths.
To prove this, we first carry out visualizations to show the biased distribution of representations learned by random sampling. Then, we delve into the gradient of InfoNCE loss and provide evidences that data bias influences the model optimization.

\noindent \textbf{Dataset bias leads to biased representation distributions.} In the upper part of Figure~\ref{fig:embeddings}, we visualize image and text embeddings learned with random sampling. Intra-dataset representations are closely gathered, while inter-dataset representations are separated. Representations are separated to three parts, \ie, VG, SBU and ``CC3M+CC12M''. Since CC3M and CC12M are composed of similar image-text pairs, representations of CC3M and CC12M are slightly overlapped. It demonstrates that the model is driven to separate representations from different datasets, and, within a training batch, the model will easily distinguish negative samples.

\noindent \textbf{Dataset bias influences the optimization of InfoNCE.} 
Since the dual-encoder model is optimized by InfoNCE loss, we first formulate the loss function and its gradient for further explorations:
\begin{scriptsize}
\begin{equation}
	\mathcal{L} = \sum_{j} \sum_{k} y_{jk} \text{log} \left( \frac{\text{exp}(s_{jk})}{\sum_{l}\text{exp}(s_{jl})} \right),  
	\nabla_{\theta} \mathcal{L} = - \sum_{j} \sum_{k} y_{jk} \nabla_{\theta} \text{log} \left( \frac{\text{exp}(s_{jk})}{\sum_{l}\text{exp}(s_{jl})} \right),
\end{equation}
\end{scriptsize}
where the similarity between the \textit{query} $j$ and the \textit{key} $k$ as $s_{jk}$. The ground-truth label corresponding to $s_{jk}$ is represented by $y_{jk} \in \{0,1\}$. We omit the temperature parameter for simplification. Then, we derive the gradient item as~\footnote{Detailed deriviations are attached in Appendix A.1.}:
\begin{scriptsize}
\begin{equation}
\begin{aligned}
	\nabla_{\theta} \mathcal{L} & = \sum_{j} \sum_{k} \left( \frac{\text{exp}(s_{jk})}{\sum_{l} \text{exp}(s_{jl})} - y_{jk} \right) \nabla_{\theta} s_{jk} \\
	& = \sum_{j} \sum_{k} \left( {\bar{p}_{jk}} - y_{jk} \right) \nabla_{\theta} s_{jk} \\
\end{aligned}	
\label{eq:gradient1}
\end{equation}
\end{scriptsize}where we could observe that the gradient term is related to the stop-gradient term $\bar{p}_{jk}$, which reflects the similarities among training samples. Negative pairs are essential for self-supervised learning methods which are based on the InfoNCE loss~\cite{cpc}. However, as suggested in Figure~\ref{fig:embeddings}, dataset bias makes the model easily separate negative samples from different data sources, resulting in the small ${\bar p}_{jk}$ and inferior gradient for negative pairs. Thus, the effectiveness of negative samples are damaged in the optimization process, especially for significant hard examples.

\begin{figure}[h]
	\centering
    \includegraphics[width=1.0\linewidth]{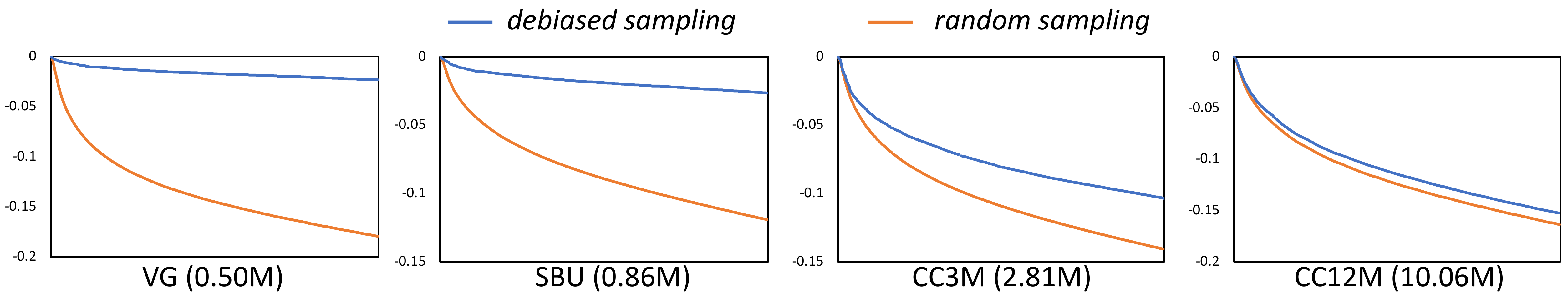}
	\caption{$\text{log}({\bar p}_{jk})$ averaged over negative pairs on different datasets and scales. The larger value contributes to the larger gradient of negative samples. } 
	\label{fig:logp}
\end{figure}

\noindent \textbf{Debiased sampling.} 
Knowledge of the dataset bias is not beneficial for downstream tasks and can be even harmful for learning essential semantic concepts. To tackle the dataset bias issue, the key factor is forcing the model to focus on helpful knowledge instead of the dataset bias. Inspired by this, we propose the debiased sampling strategy, as illustrated in Figure~\ref{fig:sampling}. Debiased sampling ensures instances within each batch come from the same dataset. For example, the first batch consists of samples from only SBU, and the second batch is composed of samples of only CC3M. Under this regularization, models are not allowed to hack the optimization by remembering the dataset bias. As shown in Figure~\ref{fig:embeddings}, the biased distributions of representations are significantly alleviated by our method, especially on the text modality. Besides, as shown in Figure~\ref{fig:logp}, it could be observed that training with debiased sampling yields larger ${\bar p}_{jk}$ of negative pairs~(on all datasets) than random sampling, \ie, debiased sampling successfully increases the effectiveness of negative samples. Figure~\ref{fig:logp} suggests that samples in smaller datasets could suffer from less effective gradient of negative samples, and our method alleviates this problem by increasing gradient of negative samples, especially for small datasets~(\ie, VG and SBU). Moreover, downstream results are remarkably improved by the debiased sampling, which will be discussed later.

\subsection{Creating Extra Data with Coin Flipping Mixup}

Intuitively, data augmentation is a ubiquitous method to create extra training data. With limited data resources, the augmentation plays an important rule in boosting performances. This part introduces a novel data augmentation method, which bring in little computational complexity but remarkably improve model performance.

\noindent \textbf{Coin flipping mixup.} To the best of our knowledge, mixup~\cite{mixup,manifold_mixup,cutmix,text_mixup,imix} are seldom investigated in the vision-language pre-training task. 
In this part, we first formulate the common mixup strategy in the dual-encoder training scheme, and reveal the label assignment dilemma when calculating contrastive loss. To solve this dilemma, we further propose a novel coin flipping mixup.

\noindent \textit{(1)~Formulations and the label assignment dilemma.}
We follow the previous works~\cite{mixup,text_mixup} by applying instance-level mixup.
Given a batch of N image-text pairs, the image and text of the $j$-th pair are denoted by $I_j$ and $T_j$, respectively. Instead of randomly mixing image-text pairs within the batch, we leverage a more efficient mixing operation for easy implementations:
\begin{scriptsize}
\begin{equation}
\begin{aligned}
	\tilde{I}_j & = \lambda * I_j + (1-\lambda) * I_{N-j}, \\
	\tilde{T}_j & = \lambda * T_j + (1-\lambda) * T_{N-j},
\end{aligned}
\label{eq:mixup_sample}
\end{equation}
\end{scriptsize}where $\tilde{I}_j$ and $\tilde{T}_j$ denote the $j$-th mixed image and text. $\lambda$ is sampled from the distribution $Beta(\alpha, \alpha)$. Therefore, the training batch after the mixing operation could be denoted by $\{(\tilde{I}_1, \tilde{T}_1), (\tilde{I}_2, \tilde{T}_2), \ldots, (\tilde{I}_N, \tilde{T}_N)\}$. 
However, we will encounter a label assignment dilemma. For instance, both $(\tilde{I}_j, \tilde{T}_j)$ and $(\tilde{I}_{N-j}, \tilde{T}_{N-j})$ are contained in the batch but interpolated by the same instances. It is not feasible to measure the target matching score between $\tilde{I}_j$ and $\tilde{T}_{N-j}$.  Particularly, the $\tilde{I}_j$ and $\tilde{T}_{N-j}$ are written as:
\begin{scriptsize}
\begin{equation}
\begin{aligned}
	\tilde{I}_j & = \lambda * I_j + (1-\lambda) * I_{N-j}, \\
	\tilde{T}_{N-j} & = (1-\lambda) * T_j + \lambda * T_{N-j},
\end{aligned}
\label{eq:mixup_dilemma}
\end{equation}
\end{scriptsize}where the similarity between $\lambda * I_j$ and $(1-\lambda) * T_j$ is not measurable based on the prior knowledge of mixup~\cite{mixup}. 

\begin{wrapfigure}{r}{.5\linewidth}
	\centering
    \includegraphics[width=1.0\linewidth]{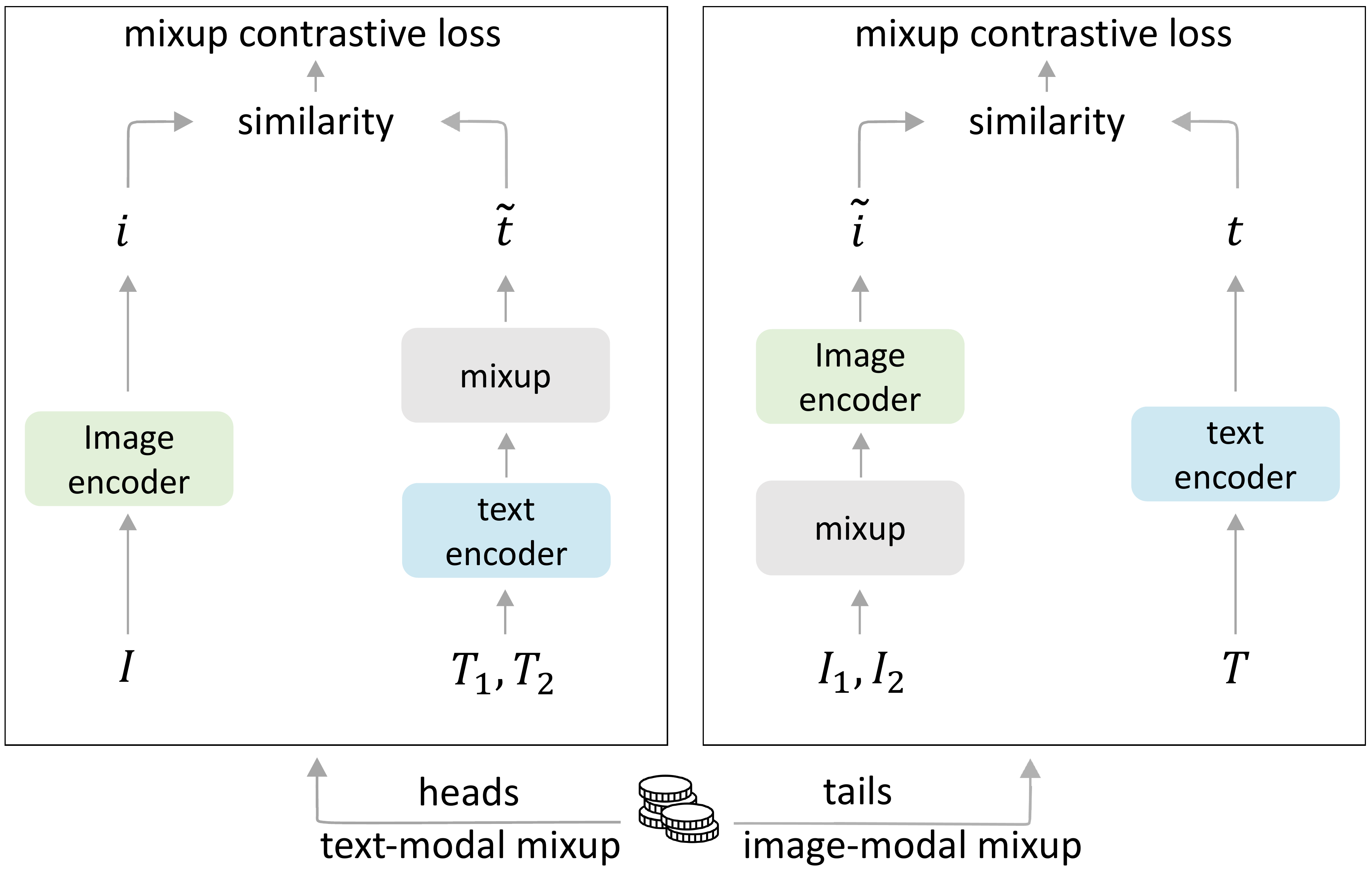}
	\caption{Illustration of our proposed coin flipping mixup. Note that manifold mixup is applied on the text modality, since we empirically observe that interpolating sparse word embeddings could lead to significant performance drop.} 
	\label{fig:coin_flipping_mixup}
\end{wrapfigure}

\noindent \textit{(2)~Coin flipping mixup.} To tackle the above problem, we propose the coin flipping mixup strategy. Briefly, mixup is applied on \textit{one of the multiple modals} in each training batch, avoiding the above label assignment dilemma. In our implementation, by uniformly sampling $\gamma$ from the range $[0, 1]$, we enable the mixup on image modal if $\gamma > 0.5$, otherwise text modal. Interestingly, as shown in Figure~\ref{fig:coin_flipping_mixup}, the strategy is similar to the coin flipping decision-making procedure, from which its name derives.

We briefly formulate the learning objective of coin flipping mixup, assuming $\gamma > 0.5$ and the mixup on image modal is enabled. In literature~\cite{clip,align}, the contrastive loss could be disentangled to image-to-text and text-to-image matching parts. Correspondingly, the mixup contrastive loss of image-to-text matching is written as:
\begin{scriptsize}
\begin{equation}
\begin{aligned}
   	\mathcal{L}_{\tilde{I}2T} & = \lambda * \left(- \frac{1}{N} \sum_{j=1}^{N} \text{log} \frac{\text{exp}(\tilde{i_j} \cdot t_j)}{\sum_{k=1}^{N}\text{exp}(\tilde{i_j} \cdot t_k)}\right ) \\ 
	& + (1-\lambda) * \left(- \frac{1}{N} \sum_{j=1}^{N} \text{log} \frac{\text{exp}(\tilde{i_j} \cdot t_{N-j})}{\sum_{k=0}^{N-1}\text{exp}(\tilde{i_j} \cdot t_{N-k})}\right ),
\label{eq:i2t}
\end{aligned}
\end{equation}
\end{scriptsize}where $\tilde{i}_{j}$ and $t_j$ respectively denote representations of the mixed image $\tilde{I}_j$ and the non-mixed text $T_j$. The text-to-image matching part shares similar formulations.

\subsection{Experiment Results and Discussions}
Main results of debiased sampling and coin flipping mixup are reported in Table~\ref{tab:data_results}. Overall speaking, both methods benefit performances on both F30K and MS-COCO. Note that these experiments only involve 14M academic data. Stacking these methods jointly contributes to $+31.2$ and $+35.9$ RSUM improvements on F30K and MS-COCO, respectively. Undoubtedly, properly leveraging limited data is of vital importance, and our methods are beneficial.

\begin{table*}[h]
\centering
\setlength{\tabcolsep}{0.8mm}{
\begin{scriptsize}
\begin{tabular}{l|ccccc|ccccc} \Xhline{2\arrayrulewidth}
         & \multicolumn{5}{c|}{MS-COCO~(5K test set)}                                                        & \multicolumn{5}{c}{Flickr30K~(1K test set)}                                                           \\ 
         & \multicolumn{2}{c}{I $\rightarrow$ T} & \multicolumn{2}{c}{T $\rightarrow$ I} &      & \multicolumn{2}{c}{I $\rightarrow$ T} & \multicolumn{2}{c}{T $\rightarrow$ I} &      \\ \hline
         & R@1                     & R@10         & R@1                     & R@10         & RSUM & R@1                     & R@10         & R@1                     & R@10         & RSUM \\
baseline  & 45.9                    & 82.8         & 35.0                    & 73.1         & 371.6 & 66.0                   & 95.1         & 58.6                    & 90.6         & 483.3  \\
+ debiased sampling  & 53.2                    & 86.4         & 36.7                    & 74.1         & 392.3 & 78.8                   & 98.2         & 61.2                    & 91.9         & 510.1  \\
+ coin flipping mixup  & 53.0                   & 87.6         & 39.6                   & 76.5         & 402.8 & 80.1                    & 98.4         & 63.7                   & 93.1         & 519.2  \\ \Xhline{2\arrayrulewidth}
\end{tabular}
\end{scriptsize}}
\caption{Results of stacking methods for training with limited data resource.}
\label{tab:data_results}
\end{table*}

\noindent \textbf{Effect of debiased sampling. }
Compared to the baseline, debiased sampling achieves consistent and remarkable improvements on all metrics, without any extra computational costs and hyper-parameters. It validates the effectiveness of our proposed debiased sampling, and debiased learning is a potential research direction in the contrastive language-image pre-training field.

\noindent \textbf{Effect of coin flipping mixup.} We set the alpha value of the beta distribution to 0.1, then apply input mixup on image modal and manifold mixup on text modal. Noticeable promotions are contributed by the coin flipping mixup, especially on text-to-image~(T2I) metrics, \ie, text-to-image Recall@1 on F30K and MS-COCO are improved by $+2.5$ and $+2.9$.

\noindent \textbf{Empirical observations on data augmentation.} (1) The cropping area of RandomResizeCrop should be in a relatively large range for covering main objects. (2) AutoAugment~\cite{autoaugment} brings in satisfactory improvements but little computational overhead. (3) Randomly masking input words advances the model performance with no cost.

\section{Training with Limited Computational Resource}
\label{sec:training}

In contrastive self-supervised learning~\cite{simclr,simclrv2}, distributed large-batch training has become a standard practice, for increasing the training batch size and providing enough negative samples. Firstly, we demonstrate the remarkable benefits of distributed large-batch training in the multi-modal pre-training task; however, it relies on considerable computational resources~(\eg, training our model with 16,384 batch size needs 128 V100 GPUs). Then, to tackle this problem, we study how to achieve comparable results with limited computational resources~(\eg, 8 V100 GPUs) by proposing the decoupled gradient accumulation. Lastly, we discuss how to accelerate the training.

\subsection{Large-Batch Training with Decoupled Gradient Accumulation}
\label{sec:gradient_reserved_gather}

\begin{table*}[h]
\centering
\setlength{\tabcolsep}{0.8mm}{
\begin{scriptsize}
\begin{tabular}{l|ccccc|ccccc} \Xhline{2\arrayrulewidth}
         & \multicolumn{5}{c|}{MS-COCO (5K test set)}                                                        & \multicolumn{5}{c}{Flickr30K (1K test set)}                                                           \\ 
         & \multicolumn{2}{c}{I $\rightarrow$ T} & \multicolumn{2}{c}{T $\rightarrow$ I} &      & \multicolumn{2}{c}{I $\rightarrow$ T} & \multicolumn{2}{c}{T $\rightarrow$ I} &      \\ \hline
         & R@1                     & R@10         & R@1                    & R@10         & RSUM & R@1                    & R@10         & R@1                    & R@10         & RSUM \\
baseline + data & 53.0                   & 87.6         & 39.6                   & 76.5         & 402.8  & 80.1                   & 98.4         & 63.7                   & 93.1         & 519.2  \\
+ gradient reserved gather    & 55.4                   & 88.7         & 42.0                   & 78.7         & 415.0 & 81.4                   & 98.2         & 66.2                  & 93.7         & 524.1  \\
$\,$  + batch = 2,048   & 56.4                   & 88.5         & 42.7                    & 79.2         & 418.0 & 81.5                    & 98.6         & 68.2                    & 93.7         & 527.5  \\
$\,$  + batch = 4,096   & 58.9                   & 89.9         & 43.8                   & 79.6         & 425.9 & 82.7                   & 98.6         & 68.7                    & 94.5         & 531.7  \\
$\,$  + batch = 8,192   & 59.0                   & 89.5         & 43.7                   & 79.5         & 424.4 & 83.1                    & 98.7         & 68.5                    & 94.6         & 531.8  \\
$\,$  + batch = 16,384  & 59.3                    & 89.6         & 44.1          & 70.4                   & 425.0 & 85.5                   & 98.5         & 69.8                    & 94.5         & 536.2  \\ \Xhline{2\arrayrulewidth}
\end{tabular}
\end{scriptsize}}
\caption{Results of distributed large-batch training. ``baseline $+$ data'' denotes the result of stacking methods proposed in Sec.~\ref{sec:data}.}
\label{tab:training}
\end{table*}

\noindent \textbf{Benefits of large-batch distributed training. }
In the practical implementation of distributed InfoNCE loss, gather operations are frequently used to collect negative samples across machines. In multi-modal scenario, the InfoNCE loss could be separated into image-to-text~(I2T) and text-to-image~(T2I) matching parts. Similar to Eqn.~(\ref{eq:gradient1}), the gradient of the I2T part is as followed~\footnote{Due to the page limit, detailed formulations are attached in Appendix A.1.}:
\begin{scriptsize}
\begin{equation}
	\nabla_{\theta} \mathcal{L}^{I2T} = \sum_{j} \sum_{k} \left( \bar{p}^{I2T}_{jk} - y^{I2T}_{jk} \right) \left( \bar{i}_{j}\nabla_{\theta}t_{k} + \bar{t}_{k}\nabla_{\theta}i_{j} \right), 
\label{eq:l_i2t}
\end{equation}
\end{scriptsize}where we place a vinculum on a value to indicate its gradient is detached. For a pair $(i_{j}, t_{k})$ from \textit{different} machines, gather operations with detaching gradients would produce the following wrong gradient on the machine of sample $j$:
\begin{scriptsize}
\begin{equation}
	\tilde{\nabla}_{\theta} \mathcal{L}_{ij}^{I2T} = \left( \bar{p}^{I2T}_{jk} - y^{I2T}_{jk} \right) \bar{t}_{k}\nabla_{\theta}i_{j}.
\end{equation} 
\end{scriptsize}Therefore, preserving gradients of gathered embeddings would provide valuable gradients. As reported in Table~\ref{tab:training}, by preserving gradients of gathered embeddings, noticeable gains are achieved within expectations. Concretely, $+4.9$ RSUM on F30K and $+12.2$ RSUM on MS-COCO are contributed by the gradient reversed gather, further supporting our derivations.

Previous works have demonstrated that self-supervised contrastive learning could significantly benefit from the large training batch size, which provides more negative examples to facilitate the model convergence~\cite{simclr}. 
To further analyze the impact of varying batch sizes on multi-modal contrastive pre-training, we scale the batch size from 1,024 to 16,384 and keep training epochs consistent.
Besides, previous works~\cite{simclr,moco} empirically showed that linearly scaling the initial learning rate is necessary for large-batch training. 
Regarding large batch experiments, up to 128 Nvidia V100 GPUs are used. As shown in Table~\ref{tab:training}, increasing the batch size from 1,024 to 16,384 leads to significant improvements on all evaluated metrics, indicating the vital importance of large-batch training. However, substantial computational resources are used for containing large batches. 

\noindent \textbf{Decoupled gradient accumulation. }
A common strategy to mimic large-batch training is the multi-step gradient accumulation. Concretely, a training iteration of a large batch is divided into several sub-iterations, and, in each sub-iteration, the batch size is relatively small. Gradients of multiple sub-iterations are individually calculated, accumulated and jointly back-propagated. It is a practical strategy in deep learning tasks; however, to mimic the large batch InfoNCE loss, the calculation process unavoidably involves embeddings from different training sub-iterations, which are, unfortunately, not accessible across sub-iterations. Therefore, the conventional multi-step gradient accumulation is not able to enlarge the effective batch size, greatly limiting final model performances. 

We propose the decoupled gradient accumulation to make large-batch contrastive learning feasible for limited resources. According to Eqn.~(\ref{eq:l_i2t}), we mathematically decouple the gradient of a large batch into two parts~\footnote{Due to the page limit, detailed derivations are attached in Appendix A.2.}:
\begin{scriptsize}
\begin{equation}
\begin{aligned}
	\nabla_{\theta} \mathcal{L} & = \nabla_{\theta} \mathcal{L}^{I2T} + \nabla_{\theta} \mathcal{L}^{T2I} \\ 
	& = \sum_{j} \nabla_{\theta} \underbrace{ \left( \sum_{k} \left( \bar{p}^{I2T}_{jk} - y^{I2T}_{jk} + \bar{p}^{T2I}_{kj} - y^{T2I}_{kj} \right) \bar{t}_{k} \right) }_{\text{stop-gradient part}} i_{j} \\
	& \quad + \sum_{k} \nabla_{\theta} \underbrace{ \left( \sum_{j} \left( \bar{p}^{I2T}_{jk} - y^{I2T}_{jk} + \bar{p}^{T2I}_{kj} - y^{T2I}_{kj} \right) \bar{i}_{j} \right) }_{\text{stop-gradient part}} t_{k},
\end{aligned}
\label{eq:ga}
\end{equation}
\end{scriptsize}where one part of gradient is only related to embeddings within each sub-iteration, and the other part only depends on stop-gradient embeddings of the large batch, which can be obtained by forwarding the large batch for an extra time. 
In this manner, we are allowed to take advantages of the large batch size by sacrificing training time.

As reported in Table~\ref{tab:ga}, it empirically shows that, by sacrificing extra 40\%--50\% training time, our gradient accumulation successfully mimics large-batch training without damaging model performances. With 8 V100 GPUs, we are not allowed to train the model with batch sizes larger than 1,024, and thus achieved performances are relatively unsatisfactory. However, our method successfully allows us to train models with large effective batch sizes 8,192 and 16,384, achieving comparable RSUM scores with only 8 V100 GPUs.

\begin{table}[h]
\centering
\setlength{\tabcolsep}{0.85mm}{
\begin{scriptsize}
\begin{tabular}{c|c|c|c|c|c|c} \Xhline{2\arrayrulewidth}
\multirow{2}{*}{batch} & DGA  & effective & \multirow{2}{*}{\# GPU} & GPU time & MS-COCO & F30K \\
                            &       step              &       batch   &    &  (hr)             & RSUM          & RSUM             \\ \hline
1,024                        & --            &   1,024 & 8 & $\sim$430              & 415.0       & 524.1                  \\ \hline
8,192                       & --           &   8,192  & 64 & --                     & 424.4  & 531.8                  \\
16,384                      & --            &    16,384 & 128 & --                  & 425.0    & 536.2                   \\ \hline 
1,024                        & 8              &   8,192 & 8 & $\sim$600             & 424.1     & 532.2                    \\ 
1,024                        & 16              &  16,384 & 8 & $\sim$680            & 425.2      & 535.9                 \\ \Xhline{2\arrayrulewidth}
\end{tabular}
\end{scriptsize}}
\caption{RSUM scores of decoupled gradient accumulation~(DGA). For training with batch 8,192 and 16,384, 64 and 128 V100 GPUs are required, respectively.}
\label{tab:ga}
\end{table}

\noindent \textbf{Stable decoupled gradient accumulation. } Note that encoders could contain modules of randomness, \eg, dropout layers are widely applied in the BERT~\cite{bert}. Thus, forwarding the same sample two times could produce different embeddings. To this end, we set the identical random seed for twice forwarding processes, eliminating the randomness and stabilizing the training. In Table~\ref{tab:stable_dga}, we provide an ablation study related to the stable training. It demonstrates that significant performance drops would be caused without considering the randomness. Forwarding the same sample for two times yields different embeddings results in the gradient in Eqn.(\ref{eq:ga}) is wrongly calculated. 
\begin{table}[h]
\centering
\setlength{\tabcolsep}{0.85mm}{
\begin{scriptsize}
\begin{tabular}{c|c|c|c|c|c|c} \Xhline{2\arrayrulewidth}
\multirow{2}{*}{batch} & DGA  & effective & \multirow{2}{*}{\# GPU} & \multirow{2}{*}{\textbf{stable}} & MS-COCO & F30K \\
                            &       step              &       batch   &    &               & RSUM          & RSUM             \\ \hline
1,024                        & 16              &  16,384 & 8 & \checkmark            & 425.2      & 535.9                 \\
1,024                        & 16              &  16,384 & 8 &    --         & 413.4      & 527.1                 \\ \Xhline{2\arrayrulewidth}
\end{tabular}
\end{scriptsize}}
\caption{Effects of stable training. ``$\checkmark$'' denotes setting the identical random seed for twice forwarding processes, and the achieved results correspond to Table~\ref{tab:ga}.}
\label{tab:stable_dga}
\end{table}

\subsection{Fast Training with TokenDrop and Auxiliary Encoders}
Thus far, all methods for better performances are elaborated. 
For real-scenario multi-modal applications, the training efficiency and model performance are equally significant for various deployment purposes. We introduce two methods on fast training for different purposes.

\noindent \textbf{TokenDrop.} Inspired by the recent work~\cite{mae}, we randomly drop a part of input pixels to speed-up the training of image encoders. Empirically, we observe that randomly masking 25\% input tokens of ViT introduces negligible performance drop, but considerably reduces training time. As shown in Table~\ref{tab:tokendrop}, enabling TokenDrop saves $\sim$30\% training time. Besides, training with TokenDrop compensates for the extra training time caused by DGA.

\begin{table}[h]
\centering
\setlength{\tabcolsep}{0.85mm}{
\begin{scriptsize}
\begin{tabular}{c|c|c|c|c|c|c} \Xhline{2\arrayrulewidth}
\multirow{2}{*}{batch} & DGA  & Token & \multirow{2}{*}{\# GPU} & GPU time & MS-COCO & F30K \\
                            &       step              &       Drop   &    &  (hr)             & RSUM          & RSUM             \\ \hline
1,024                        & 16              &   -- & 8 & $\sim$680             & 425.2     & 535.9                    \\ 
1,024                        & 16              &  \checkmark & 8 & $\sim$\textbf{470}            & 424.8      & 535.5                 \\ \Xhline{2\arrayrulewidth}
\end{tabular}
\end{scriptsize}}
\caption{Training time saved by TokenDrop. }
\label{tab:tokendrop}
\end{table}

\noindent \textbf{Auxiliary Encoders.}
Assuming that we have trained a model with heavy encoders, we investigate how to fast obtain lightweight encoders with auxiliary heavy ones. Since the training of a dual-encoder model is driven by the InfoNCE loss, embeddings yielded by either encoder are regarded as the ``learning target'' of the other side. Thus, enlarging either encoder's capacity would contribute to more reliable and discriminative embeddings. Assuming that we have trained a dual-encoder model with heavy encoders, \eg, ViT-B/16 and BERT-Base, we can replace one of them to a lightweight one and re-train it with the guidance of the other one in a distillation manner~\cite{kd,dkd}. 
For instance, we change the image encoder from ViT-B/16 to ViT-B/32, and then re-train it with the BERT-Base being frozen. 
With the guidance of the frozen encoder, the training process of the replaced encoder could be greatly accelerated, as reported in Table~\ref{tab:auxiliary}. 

\begin{table}[h]
\centering
\setlength{\tabcolsep}{0.8mm}{
\begin{scriptsize}
\begin{tabular}{c|cc|c|c|c} \Xhline{2\arrayrulewidth}
training & \multicolumn{2}{c|}{encoder} & GPU & MS-COCO & F30K \\
method                            &    image                 &    text   & time (hr)                     & RSUM          & RSUM        \\ \hline
\multirow{2}{*}{auxiliary}                       & ViT-B/16                  & BERT-B                      & --  & 402.8 & 519.1                   \\
                        & ViT-B/32                   & BERT-B$\spadesuit$      &         $\sim$\textbf{110}       & 381.2 & 493.9                   \\ \hline
baseline                        & ViT-B/32                   & BERT-B &  $\sim$240                 & 379.5   & 494.1                  \\ \hline   \Xhline{2\arrayrulewidth}
\end{tabular}
\end{scriptsize}}
\caption{Training time saved by the auxiliary encoder method. ``$\spadesuit$'' symbol denotes the model is frozen.}
\label{tab:auxiliary}
\end{table}

\section{Comparisons with SOTA Methods}
\label{sec:sota}

In this section, we focus on assessing the pre-training performances with cross-modal retrieval tasks, in both zero-shot and fine-tuned settings~\cite{clip,align}. We name our method as ``ZeroVL'', where ``Zero'' means the motivation for designing a strong baseline with limited resources.

\subsection{Zero-Shot Cross-Modal Retrieval}

\begin{table*}[h]
\centering
\setlength{\tabcolsep}{0.25mm}{
\begin{tiny}
\begin{tabular}{l|cc|c|c|c|ccccc|ccccc} \Xhline{2\arrayrulewidth}
 & \multicolumn{2}{c|}{computation}  &  \multirow{2}{*}{data} &  input   & batch  & \multicolumn{5}{c|}{MS-COCO (5K test set)}                                                        & \multicolumn{5}{c}{Flickr30K (1K test set)}                                                           \\ 
 & device & count &  & size  & size  & \multicolumn{2}{c}{I $\rightarrow$ T} & \multicolumn{2}{c}{T $\rightarrow$ I} &      & \multicolumn{2}{c}{I $\rightarrow$ T} & \multicolumn{2}{c}{T $\rightarrow$ I} &      \\ \hline
 \textbf{\textit{zero-shot}}  &  & &  &  &  & R@1             & R@10         & R@1                  & R@10         & RSUM & R@1           & R@10         & R@1     & R@10         & RSUM \\
CLIP & V100 & 256 &  400M & 336 & 32,768 & 58.4              & 88.1         & 37.8             & 72.2         & 400.2 & 88.0             & 99.4         & 68.7             & 95.2         & 540.6  \\
ALIGN & TPU$_{\text{v3}}$ & 1,024 & 1800M & 289 & 16,384 & 58.6                & 89.7         & 45.6                & 78.6         & 425.3 & \textbf{88.6}               & \textbf{99.7}         & \textbf{75.7}            & \textbf{96.8}         & \textbf{553.3}  \\  \hline
baseline & V100 & 8 & 14M & 224 & 1,024 & 45.9                    & 82.8         & 35.0                    & 73.1         & 371.6 & 66.0                   & 95.1         & 58.6                    & 90.6         & 483.3   \\
CLIP~(our impl.) & V100 & 8 &  14M & 224 & 1,024 & 51.0              & 85.5         & 38.2             & 75.5         & 392.5 & 80.9             & 97.8         & 63.8             & 92.4         & 518.4  \\

CLIP~(our impl.) & V100 & 128 &  14M & 224 & 16,384 & 57.7              & 88.7         & 41.6             & 77.8         & 416.0 & 83.1             & 98.3         & 67.2             & 93.9         & 527.3  \\
ZeroVL~(ours) & V100 & 8 & 14M & 224 & 16,384 & 59.3                    & 89.6         & 44.1                    & 79.5         & 425.0 & 85.5                    & 98.5         & 69.8                    & 94.5         & 536.2  \\ \hline
ZeroVL$\dagger$~(ours)  & V100 & 8 & 100M & 224 & 16,384 & \textbf{64.0}                    & \textbf{91.4}         & \textbf{47.3}                    & \textbf{81.1}         & \textbf{442.1} & 88.0                   & 99.2         & 73.5                    & 95.7         & 546.5  \\ 
\Xhline{2\arrayrulewidth}
\end{tabular}
\end{tiny}}
\caption{\textit{{Z}ero-shot} cross-modal retrieval results. ``baseline'' is the naive baseline in Sec.~\ref{sec:dataset}. ``$\dagger$'' means training with the 100M web data. }
\label{tab:zs_sota}
\end{table*}

\noindent \textbf{Setup.} Training implementation details are as followed. On the ground of baseline settings~(\eg, learning rate, training epoch, and weight decay) introduced in Sec.~\ref{subsec:baseline_setting}, we stack all proposed methods, \ie, debaised sampling, coin flipping mixup, and decoupled gradient accumulation. For reproducibility, we mainly benchmark with publicly accessible academic datasets. For fair comparisons, we re-implement CLIP with 14M data to validate the performance drop caused by limited resources.
Besides, CLIP and ALIGN respectively collect 400M and 1.8B image-text pairs from the web. Due to training datasets of CLIP and ALIGN are not available, we also collect $\sim$100M web image-text pairs for validating the effectiveness of our method on large-scale data. 

\noindent \textbf{Main results.}
In Table~\ref{tab:zs_sota}, on both F30K and MS-COCO datasets, we achieve competitive results on the basis of 14M academic publicly accessible data and 8 V100 GPUs. It is worth mentioning that our ZeroVL already exceeds CLIP on the MS-COCO dataset in both image-to-text~(I2T) and text-to-image~(T2I) metrics, \eg, our I2T R@1 and T21 R@1 surpass CLIP by $+0.9$ and $+6.3$, respectively. Results of our implemented CLIP further validate the contribution of our efforts, \ie, the performance of cross-modal retrieval would be greatly suppressed if the resources were greatly limited. 
In addition, although our collected 100M web images are much less than those of CLIP and ALIGN, ZeroVL still successfully outperforms CLIP trained with 400M data and ALIGN trained with 1.8B data on MS-COCO. On F30K, we perform slightly worse than ALIGN but better than CLIP, which can result from the domain of ALIGN's data is larger than ours.

\noindent \textbf{Resource costs.} For computational resources, training CLIP requires 256 V100 GPUs, and training ALIGN needs 1,024 Could TPUv3 cores. Experiments in Table~\ref{tab:zs_sota} involve 8 V100 32GB GPUs. For data resources, we mainly benchmark on 14M publicly accessible academic datasets to guarantee the reproducibility. Experiments of 100M web data demonstrate that our method is still effective on large-scale data, \ie, our method fits in different data scales without tuning hyper-parameters. Additionally, only 2.4 days are required for training ZeroVL with 8 V100 and 14M academic data, which could be friendly to the most researchers.

\subsection{Fine-Tuned Cross-Modal Retrieval}

\begin{table*}[h]
\centering
\setlength{\tabcolsep}{0.4mm}{
\begin{tiny}
\begin{tabular}{l|c|cc|ccccc|ccccc} \Xhline{2\arrayrulewidth}
 & input  &  \multicolumn{2}{c|}{encoder}     & \multicolumn{5}{c|}{MS-COCO (5K test set)}                                                        & \multicolumn{5}{c}{Flickr30K (1K test set)}                                                           \\ 
 & size &   image~(I) & text~(T)    & \multicolumn{2}{c}{I $\rightarrow$ T} & \multicolumn{2}{c}{T $\rightarrow$ I} &      & \multicolumn{2}{c}{I $\rightarrow$ T} & \multicolumn{2}{c}{T $\rightarrow$ I} &      \\ \hline
\textbf{\textit{fine-tuned}} &   &   &   & R@1                   & R@10         & R@1                     & R@10         & RSUM & R@1                     & R@10         & R@1                      & R@10         & RSUM \\
VSE++ & 512 & RX101* & BERT-B & 57.9                   & 92.8         & 44.9                    & 84.0         & 439.2 & 80.9                   & 98.9         & 65.2                    & 93.7         & 524.8 \\
GPO & 512 & RX101* & BERT-B & 68.1                   & 95.2         & 52.7                    & 88.3         & 474.8 & 88.7                   & 99.8         & 76.1                    & 97.1         & 555.1 \\
ALIGN & 289 & EffNet-L2* & BERT-L & 77.0                   & 96.9         & \textbf{59.9}                   & 89.8         & 500.4  & \textbf{95.3}                  & \textbf{100.0}         & \textbf{84.9}                   & \textbf{98.6}         & \textbf{576.0} \\ \hline
baseline & 224 & ViT-B/16 & BERT-B & 69.1                   & 94.8         & 51.9                    & 86.8         & 471.9 & 90.1                   & 99.1         & 75.1                   & 96.6         & 553.0  \\
CLIP~(our impl. 8V100) & 224 & ViT-B/16 & BERT-B & 69.9                    & 94.9         & 52.5                    & 87.0         & 473.8 & 90.4                   & 99.2         & 75.6                  & 96.5         & 554.1  \\
CLIP~(our impl. 128V100) & 224 & ViT-B/16 & BERT-B & 71.7                    & 95.8         & 54.0                    & 88.1         & 481.3 & 91.1                   & 99.5         & 78.5                  & 97.7         & 560.7  \\ 
ZeroVL~(ours) & 224 & ViT-B/16 & BERT-B & 72.9                    & 95.9         & 55.1                    & 88.6         & 485.0 & 91.7                   & 99.5         & 79.2                  & 97.1         & 561.6  \\ \hline
ZeroVL$\dagger$~(ours)  & 288 & ViT-B/16 & BERT-B & \textbf{77.2}                   & \textbf{97.1}         & 59.3                   & \textbf{90.2}         & \textbf{500.5} & 95.0                   & 100.0          & 83.7           &  98.6      & 573.6  \\ 
 \Xhline{2\arrayrulewidth}
\end{tabular}
\end{tiny}}
\caption{\textit{{F}ine-tuned} cross-modal retrieval results of representative dual-encoder methods. ``RX101*'' correspond to the ResNeXt-101 model pre-trained on Instagram-1B~\cite{wsl}. ``EffNet-L2*'' denotes the large CNN model EfficientNet-L2~\cite{efficientnet,efficientnetl2}. ``$\dagger$'' denotes pre-training with the 100M web data. }
\label{tab:sota}
\end{table*}

\noindent \textbf{Setup.} After the pre-training phase, we fine-tune the model on downstream datasets F30K and MS-COCO. Fine-tuning hyper-parameters are identical to pre-training's, except the initial learning rate, training epoch, and batch size. The is learning rate is set to 1e-5. For F30K and MS-COCO, we optimize the model for 1K and 5K steps. Batch size is set to 2,048. Similar to zero-shot experiments, we also provide fine-tuning results with both 14M and 100M data.

\noindent \textbf{Main results.} In Table~\ref{tab:sota}, with 14M academic pre-training data, we successfully outperforms state-of-the-art in-domain training method VSE++~\cite{vsepp} and GPO~\cite{gpo}. It is worth mentioning that GPO also involves large-scale pre-training on the image modal, \ie, weakly supervised pre-training with the Instagram-1B dataset~\cite{wsl}. Compared with GPO, ZeroVL can achieve better results with the more efficient image encoder and smaller training input size, strongly supporting the effectiveness of our pre-training method. For experiments with 100M web data, it is worth noting that ALIGN uses (1) significantly more pre-training data, (2) heavier image and text encoders, and~(3) larger pre-training resolutions than our method. Nevertheless, similar to results in zero-shot, we still achieve comparable results to ALIGN.

\subsection{Linear Probing}

\begin{table}[h]
\centering
\setlength{\tabcolsep}{0.8mm}{
\begin{scriptsize}
\begin{tabular}{c|cc|c|c|c|c|c|c} \Xhline{2\arrayrulewidth}
\multirow{3}{*}{} &  \multicolumn{5}{c|}{pre-training} & \multicolumn{3}{c}{linear probing} \\ \cline{2-9}
 & \multicolumn{2}{c|}{computation} &  \multirow{2}{*}{data} & input & batch & \multirow{2}{*}{backbone (\#params)} & input & top-1 \\ 
                            &    device                 &    count   &                     & size    & size   &   & size & accuracy        \\ \hline 
CLIP                       & V100                  & 256                      & 400M  & 224 & 32,768  & ViT-B/16~(87M) & 224 & 80.2                   \\
ALIGN                        & TPU$_{\text{v3}}$                   & 1,024 &  1800M                 & 289   & 16,384    & EffNet-L2~(480M) & 600     & 85.5          \\ \hline   
CLIP~(our impl.)                       & V100                  & 8                      & 14M  & 224 & 1,024 & ViT-B/16~(87M) & 224 & 75.9                   \\
CLIP~(our impl.)                       & V100                  & 128                      & 14M  & 224 & 16,384 & ViT-B/16~(87M) & 224 & 80.0                   \\  
ZeroVL~(ours)                       & V100                  & 8                      & 14M  & 224 & 16,384 & ViT-B/16~(87M) & 224 & \textbf{80.9}                   \\ 
\Xhline{2\arrayrulewidth}
\end{tabular}
\end{scriptsize}}
\caption{Linear probing results on ImageNet-1K.}
\label{tab:linear}
\end{table}

\noindent \textbf{Setup.} Following~\cite{clip,align}, we conduct the linear probing task on ImageNet-1K~\cite{imagenet} after the pre-training phase. The batch size is set to 16,384 and learning rate is set to 6.4. We optimize the model for 90 epochs with the LARS optimizer~\cite{lars}, and weight decay is set to 0. To reveal the effects of our proposed methods on linear probing, we also evaluate the re-implemented CLIP as mentioned above.

\noindent \textbf{Main results.} In Table~\ref{tab:linear}, ZeroVL out-performs CLIP by 0.7\%. However, similar to fine-tuned cross-modal retrieval, ALIGN achieves better results than ZeroVL based on heavier pre-training costs, larger model capacity, and larger image resolutions. Moreover, there are two observations on re-implemented CLIP. Firstly, we observe that training with limited computation resource~(8 V100) achieves unsatisfactory top-1 accuracy 75.9\%. Secondly, training CLIP with rich computation resource~(128 V100) greatly improves the accuracy to 80.0\%. The differences between ZeroVL and re-implemented CLIP~(with 128 V100) are methods proposed in Sec.~\ref{sec:data}, validating the effectiveness of our proposed debiased sampling and coin flipping mixup. Benefits of our methods for cutting down the heavy resources dependency are further confirmed.

\section{Conclusion}
This work provides a training guideline for conducting dual-encoder multi-modal representation contrastive learning with limited resources. The proposed methods significantly lower computational resources, while still achieving good performance to be applied in other vision-language downstream tasks. With only 14M publicly accessible academic datasets and 8 V100 GPUs, we provide a reproducible strong baseline. In addition, we achieve comparable or superior performances than state-of-the-art methods with 100M web data. We hope our training pipeline and benchmark will be useful for future researches in the multi-modal representation learning field and benefit the community.

\newcounter{alphasect}
\def\alphainsection{0}

\let\oldsection=\section
\def\section{%
  \ifnum\alphainsection=1%
    \addtocounter{alphasect}{1}
  \fi%
\oldsection}%

\renewcommand\thesection{%
 \ifnum\alphainsection=1%
   \Alph{alphasect}%
 \else
   \arabic{section}%
 \fi%
}%

\newenvironment{alphasection}{%
  \ifnum\alphainsection=1%
    \errhelp={Let other blocks end at the beginning of the next block.}
    \errmessage{Nested Alpha section not allowed}
  \fi%
  \setcounter{alphasect}{0}
  \def\alphainsection{1}
}{%
  \setcounter{alphasect}{0}
  \def\alphainsection{0}
}%

\begin{alphasection}
\section{Appendix}
The appendix is composed of 9 parts. In Sec.~\ref{subsec:gradient}, we discuss the gradient of multi-modal contrastive loss. In Sec.~\ref{subsec:gradient_accumulation}, we elaborate the derivations and implementations of decoupled gradient accumulation. In Sec.~\ref{subsec:mixup_loss}, we introduce the detailed calculation of coin flipping mixup loss. In Sec~\ref{subsec:sequential}, we explore another sampling strategy related to our proposed debiased sampling. 
In Sec~\ref{subsec:dataset_bias}, we show that debiased sampling tackles various kinds of data bias. In Sec~\ref{subsec:ds_on_one_dataset}, we show that debiased sampling works well on a single dataset.
In Sec~\ref{subsec:linear_prob}, we provide linear probing results on more datasets. In Sec.~\ref{subsec:datasets}, we detail open-source and web pre-training data. In Sec.~\ref{subsec:training_detail}, we provide training details for reproducing our strong baseline.

\subsection{Gradient of Multi-Modal Contrastive Loss}
\label{subsec:gradient}

\noindent \textbf{Formulating gradients of contrastive loss.}
Within each training batch, define the similarity between the \textit{query} $j$ and the \textit{key} $k$ as $s_{jk}$. The ground-truth label corresponding to $s_{jk}$ is represented by $y_{jk} \in \{0,1\}$. 
The contrastive loss can be formulated as:
\begin{small}
\begin{equation}
	\mathcal{L} = \sum_{j} \sum_{k} y_{jk} \text{log} \left( \frac{\text{exp}(s_{jk})}{\sum_{l}\text{exp}(s_{jl})} \right),
\end{equation}
\end{small}where the temperature parameter is omitted for simplification.
Then, the gradient of the popular contrastive loss could be written as:
\begin{small}
\begin{equation}
\begin{aligned}
	\nabla_{\theta} \mathcal{L} & = - \sum_{j} \sum_{k} y_{jk} \nabla_{\theta} \text{log} \left( \frac{\text{exp}(s_{jk})}{\sum_{l}\text{exp}(s_{jl})} \right) \\
	& = - \sum_{j} \sum_{k} y_{jk} \left( \nabla_{\theta} s_{jk} - \nabla_{\theta}  \text{log}\sum_{l}\text{exp}(s_{jl}) \right) \\
	& = - \sum_{j} \sum_{k} y_{jk} \left( \nabla_{\theta} s_{jk} -  \frac{1}{\sum_{l} \text{exp}(s_{jl})} \nabla_{\theta}  \sum_{l} \text{exp}(s_{jl}) \right) \\
	& = - \sum_{j} \sum_{k} y_{jk} \left( \nabla_{\theta} s_{jk} -  \sum_{l}\frac{\text{exp}(s_{jl})}{\sum_{m} \text{exp}(s_{jm})} \nabla_{\theta}  s_{jl} \right) \\
	& = - \sum_{j} \sum_{k} y_{jk} \left( \nabla_{\theta} s_{jk} -  \sum_{l}\bar{p}_{jl} \nabla_{\theta}  s_{jl} \right) \\
	& = - \sum_{j} \sum_{k} y_{jk}  \nabla_{\theta} s_{jk} + \sum_{j} \sum_{k} y_{jk} \sum_{l}\bar{p}_{jl} \nabla_{\theta}  s_{jl},
\end{aligned}	
\label{eq:gradient1}
\end{equation}
\end{small}where we place a vinculum on a value to indicate its gradient is detached. Due to $\sum_{k}y_{jk}=1$, we rewrite Eqn.(\ref{eq:gradient1}) as:
\begin{small}
\begin{equation}
\begin{aligned}
	\nabla_{\theta} \mathcal{L} & = - \sum_{j} \sum_{k} y_{jk} \nabla_{\theta} s_{jk} +  \sum_{j}\sum_{l}\bar{p}_{jl} \nabla_{\theta}  s_{jl}  \\
	& = - \sum_{j} \sum_{k} y_{jk} \nabla_{\theta} s_{jk} +  \sum_{j}\sum_{k}\bar{p}_{jk} \nabla_{\theta}  s_{jk} \\ 
	& = \sum_{j} \sum_{k} \left( \bar{p}_{jk} - y_{jk} \right) \nabla_{\theta} s_{jk} \\
	& = \sum_{j} \sum_{k} \left( \bar{p}_{jk} - y_{jk} \right) \left( \bar{x}_{j}\nabla_{\theta}x_{k} + \bar{x}_{k}\nabla_{\theta}x_{j} \right),
\end{aligned}	
\label{eq:gradient}
\end{equation}
\end{small}where $x_{j}$ and $x_{k}$ are embeddings of sample $j$ and $k$.
Regarding the sample $j$ as the \textit{query}, its gradient comes to $\sum_{k} \left( \bar{p}_{jk} - y_{jk} \right) \left( \bar{x}_{j}\nabla_{\theta}x_{k} + \bar{x}_{k}\nabla_{\theta}x_{j} \right)$. If sample $j$ and $k$ are from different machines, detaching gradients makes the term $\bar{x}_{j}\nabla_{\theta}x_{k}$ to $0$, since $x_k$ serves as a constant term in the gradient calculation process.

%
%

\vspace{5pt}
\noindent \textbf{Detaching gradients in multi-modal contrastive loss.}
Subsequently, we study the gradients of multi-modal contrastive loss. 
We start with minor notation adjustments to cater for the multi-modal setting. 
The calculation of multi-modal contrastive loss can be divided into image-to-text~(I2T) matching and text-to-image~(T2I) matching parts. Gradients of I2T and T2I matching losses are:
\begin{small}
\begin{equation}
	\nabla_{\theta} \mathcal{L}^{I2T} = \sum_{j} \sum_{k} \left( \bar{p}^{I2T}_{jk} - y^{I2T}_{jk} \right) \left( \bar{i}_{j}\nabla_{\theta}t_{k} + \bar{t}_{k}\nabla_{\theta}i_{j} \right), 
\label{eq:l_i2t}
\end{equation}
\begin{equation}
	\nabla_{\theta} \mathcal{L}^{T2I} = \sum_{j} \sum_{k} \left( \bar{p}^{T2I}_{jk} - y^{T2I}_{jk} \right) \left( \bar{t}_{j}\nabla_{\theta}i_{k} + \bar{i}_{k}\nabla_{\theta}t_{j} \right),
\label{eq:l_t2i}
\end{equation}
\end{small}where $i$ and $t$ represent image and text embeddings. For pairs $(i_{j}, t_{k})$ and $(t_{j}, i_{k})$ from \textit{different} machines, gather operations with detaching gradients would produce the following gradients on the machine of $j$:
\begin{small}
\begin{equation}
	\tilde{\nabla}_{\theta} \mathcal{L}^{I2T} =  \left( \bar{p}^{I2T}_{jk} - y^{I2T}_{jk} \right) \bar{t}_{k}\nabla_{\theta}i_{j}, 
\end{equation} 
\end{small}and the gradient on $k$'s machine:
\begin{small}
\begin{equation}
	\tilde{\nabla}_{\theta} \mathcal{L}^{T2I} =  \left( \bar{p}^{T2I}_{kj} - y^{T2I}_{kj} \right) \bar{i}_{j}\nabla_{\theta}t_{k}.
\end{equation}
\end{small}
We add a tilde symbol on the gradient $\tilde{\nabla}_{\theta} \mathcal{L}$, indicating the calculation involves detaching gradients. Then, we have:
\begin{small}
\begin{equation}
	\nabla_{\theta} \mathcal{L}^{I2T} + \nabla_{\theta} \mathcal{L}^{T2I} \neq \left( \tilde{\nabla}_{\theta} \mathcal{L}^{I2T} + \tilde{\nabla}_{\theta} \mathcal{L}^{T2I}  \right).
\end{equation}
\end{small}Mathematically, detaching gradients in multi-modal contrastive loss yields incorrect gradients. Experiments in Sec.~4.1 of the manuscript prove that gradient reserved gather operations are beneficial in multi-modal contrastive learning. 

\subsection{Decoupled Gradient Accumulation}
\label{subsec:gradient_accumulation}

\noindent \textbf{Decoupling the gradient of multi-modal contrastive loss. }
Inspired by a technical report\footnote{\href{https://spaces.ac.cn/archives/8471}{https://spaces.ac.cn/archives/8471}} which decouples the gradient of single-modal contrastive loss, we further generalize it to the multi-modal scenario.
According to Eqn.(\ref{eq:l_i2t}) and~(\ref{eq:l_t2i}), we have:
\begin{small}
\begin{equation}
\begin{aligned}
	\nabla_{\theta} \mathcal{L}^{I2T} & = \sum_{j} \sum_{k} \left( \bar{p}^{I2T}_{jk} - y^{I2T}_{jk} \right) \left( \bar{i}_{j}\nabla_{\theta}t_{k} + \bar{t}_{k}\nabla_{\theta}i_{j} \right) \\ 
	 & =   \sum_{k} \nabla_{\theta} \left( \sum_{j} \left( \bar{p}^{I2T}_{jk} - y^{I2T}_{jk} \right) \bar{i}_{j} \right) t_{k} \\
	 & +  \sum_{j} \nabla_{\theta} \left( \sum_{k} \left( \bar{p}^{I2T}_{jk} - y^{I2T}_{jk} \right) \bar{t}_{k} \right) i_{j}
\end{aligned}
\end{equation}
\end{small}
\begin{small}
\begin{equation}
\begin{aligned}
	\nabla_{\theta} \mathcal{L}^{T2I} & = \sum_{j} \sum_{k} \left( \bar{p}^{T2I}_{jk} - y^{T2I}_{jk} \right) \left( \bar{t}_{j}\nabla_{\theta}i_{k} + \bar{i}_{k}\nabla_{\theta}t_{j} \right)  \\
	& =   \sum_{k} \nabla_{\theta} \left( \sum_{j} \left( \bar{p}^{T2I}_{jk} - y^{T2I}_{jk} \right) \bar{t}_{j} \right) i_{k} \\
    & \quad +  \sum_{j} \nabla_{\theta} \left( \sum_{k} \left( \bar{p}^{T2I}_{jk} - y^{T2I}_{jk} \right) \bar{i}_{k} \right) t_{j} \\ 
	& =  \sum_{j} \nabla_{\theta} \left( \sum_{k} \left( \bar{p}^{T2I}_{kj} - y^{T2I}_{kj} \right) \bar{t}_{k} \right) i_{j} \\
	& \quad +   \sum_{k} \nabla_{\theta} \left( \sum_{j} \left( \bar{p}^{T2I}_{kj} - y^{T2I}_{kj} \right) \bar{i}_{j} \right) t_{k} .
\end{aligned}
\end{equation}
\end{small}Then, the total gradient can be written as:
\begin{small}
\begin{equation}
\begin{aligned}
	\nabla_{\theta} \mathcal{L} & = \nabla_{\theta} \mathcal{L}^{I2T} + \nabla_{\theta} \mathcal{L}^{T2I} \\ 
	& = \sum_{j} \nabla_{\theta} \left( \sum_{k} \left( \bar{p}^{I2T}_{jk} - y^{I2T}_{jk} + \bar{p}^{T2I}_{kj} - y^{T2I}_{kj} \right) \bar{t}_{k} \right) i_{j} \\
	& \quad + \sum_{k} \nabla_{\theta} \left( \sum_{j} \left( \bar{p}^{I2T}_{jk} - y^{I2T}_{jk} + \bar{p}^{T2I}_{kj} - y^{T2I}_{kj} \right) \bar{i}_{j} \right) t_{k}.
\end{aligned}
\label{eq:ga}
\end{equation}
\end{small}As suggested in Eqn.(\ref{eq:ga}), we mathematically decouple the gradient into two parts. One part of gradient is only related to stop-gradient embeddings~($\bar{t}_k$ and $\bar{i}_j$), and the other part only depends on embeddings with gradients~($t_k$ and $i_j$).

\vspace{5pt}
\noindent \textbf{Implementation of decoupled gradient accumulation. } In the conventional multi-step gradient accumulation, we are not allowed to obtain embeddings (with gradients) from different training sub-iterations. However, we can cache stop-gradient embeddings of the large batch, and then calculate the correct gradient with Eqn.(\ref{eq:ga}) in each sub-iteration. With forwarding the large batch and caching stop-gradient embeddings, our decoupled gradient accumulation can accurately produce the gradient produced by large-batch training.

\vspace{5pt}
\noindent \textbf{Complete pseudo code in a PyTorch-like style. }
In Sec.~4.2 of the manuscript, we provide a simplified pseudo code of decoupled gradient accumulation. In Algorithm~\ref{algo:dga}, we provide a detailed and complete pseudo code of decoupled gradient accumulation for better understanding. In the implementation of previous methods~\cite{clip,align}, the temperature of contrastive loss is learnable. Thus, in the implementation of decoupled gradient accumulation, we need consider the gradient of the temperature variable. As shown in Algorithm~\ref{algo:dga}, we detach the gradient of temperature~(with $\texttt{torch.no\_grad}$) for forwarding the large batch, and then calculate the gradient of temperature in each sub-iteration. Besides, a square-root should be applied on the value of temperature for correctly calculating the scale of temperature. Note that encoders could contain modules of randomness, \eg, dropout layers are widely applied in the BERT~\cite{bert}. Thus, forwarding the same sample two times could produce different embeddings. To this end, we set the identical random seed for twice forwarding processes, eliminating the randomness and stabilizing the training.

\begin{algorithm*}
\caption{Pseudo code in a PyTorch-like style.}
\label{algo:dga}
\scriptsize
\begin{alltt}

\color{OliveGreen}
# stable_random_seed: random seed generated by time.time()
# temp: temperature
\color{Black}
\color{OliveGreen}# fix dropout with fixed random seed \color{Black}
\color{magenta}setup_seed\color{Black}(random_seed)

\color{magenta}with\color{Black} torch.no_grad():
    \color{OliveGreen}# stop-grad forward \color{Black}
    img_emb_local, text_emb_local = [], []
    \color{magenta}for\color{Black} _idx_l \color{magenta}in\color{Black} range(0, bs, bs_train):
        _data_batch = data_batch[_idx_l: _idx_l + bs_train]
        _img_embs, _text_embs, temp = model(_data_batch)
        img_emb_local.append(_img_embs)
        text_emb_local.append(_text_embs)
    
    \color{OliveGreen}# concatenate embeddings of each GPU \color{Black}
    img_emb_local = torch.cat(img_emb_local, dim = 0)
    text_emb_local = torch.cat(text_emb_local, dim = 0)
    
    \color{OliveGreen}# gather embeddings of all GPUs \color{Black}
    img_emb_global = torch.cat(gather(img_emb_local), dim = 0)
    text_emb_global = torch.cat(gather(text_emb_local), dim = 0)

    \color{OliveGreen}# calculate cosine similarity \color{Black}
    sim_i2t_nm = img_emb_global @ text_emb_local.T / temp
    sim_i2t_mn = img_emb_local @ text_emb_global.T / temp

    \color{OliveGreen}# calculate the normalized factor in softmax function \color{Black}
    sim_i2t_esum_local = torch.sum(torch.exp(sim_i2t_mn), dim = 1)
    sim_t2i_esum_local = torch.sum(torch.exp(sim_i2t_nm.T), dim = 1)
    sim_i2t_esum = torch.cat(gather(sim_i2t_esum_local), 0).unsqueeze(dim = 1)
    sim_t2i_esum = torch.cat(gather(sim_t2i_esum_local), 0).unsqueeze(dim = 1)
    
    \color{OliveGreen}# calculate the probability matrix \color{Black}
    prob_i2t_mn = torch.exp(sim_i2t_mn) / sim_i2t_esum[bs * rank: bs * (rank + 1), :]
    prob_t2i_nm = torch.exp(sim_i2t_mn.T) / sim_t2i_esum
    prob_i2t_nm = torch.exp(sim_i2t_nm) / sim_i2t_esum
    prob_t2i_mn = torch.exp(sim_i2t_nm.T) / sim_t2i_esum[bs * rank: bs * (rank + 1), :]
    
    left_I = (prob_i2t_mn + prob_t2i_nm.T) @ text_emb_global - text_emb_local * 2 
    left_I /= torch.sqrt(temp)
    left_T = (prob_i2t_nm.T + prob_t2i_mn) @ img_emb_global - img_emb_local * 2 
    left_T /= torch.sqrt(temp)

\color{OliveGreen}# Fix dropout with fixed random seed \color{Black}
\color{magenta}setup_seed\color{Black}(random_seed)

\color{OliveGreen}# forward with grad \color{Black}
\color{magenta}for\color{Black} _idx_l \color{magenta}in\color{Black} range(0, bs, bs_train):
    _left_I = left_I[_idx_l: _idx_l + bs_train]
    _left_T = left_T[_idx_l: _idx_l + bs_train]
    _data_batch = data_batch[_idx_l: _idx_l + bs_train]

    _img_embs, _text_embs, temp = model(_data_batch)
    
    loss_i = _left_I * _img_embs
    loss_t = _left_T * _text_embs
    \color{OliveGreen}# loss corresponds to Eqn.(\ref{eq:ga}) \color{Black}
    loss = (loss_i + loss_t).sum() / 2 / bs / torch.sqrt(temp)
    \color{OliveGreen}# backward propagation \color{Black}
    loss.backward()

\color{OliveGreen}# update model parameters \color{Black}
update(model.param)

\end{alltt}
\end{algorithm*}

\subsection{Coin Flipping Mixup Loss}
\label{subsec:mixup_loss}
We detail the coin flipping mixup loss function by following notations defined in Sec. 3.2 of the manuscript. We first define a batch $\{(I_1, T_1), (I_2, T_2), \ldots, (I_N, T_N)\}$ of N image-text pairs. Then, we uniformly sample a $\gamma$ from the range $[0,1]$.

\vspace{5pt}
\noindent \textbf{$\gamma > 0.5$:} We apply the mixup on the images, and the mixed batch can be denoted as $\{(\tilde{I}_1, T_1), (\tilde{I}_2, T_2), \ldots, (\tilde{I}_N, T_N)\}$. The image-to-text matching part can be formulated as:
\begin{small}
\begin{equation}
\begin{aligned}
	\mathcal{L}_{\tilde{I}2T} & = \lambda * \left(- \frac{1}{N} \sum_{j=1}^{N} \text{log} \frac{\text{exp}(\tilde{i_j} \cdot t_j / \tau)}{\sum_{k=1}^{N}\text{exp}(\tilde{i_j} \cdot t_k / \tau)}\right ) \\ 
	& + (1-\lambda) * \left(- \frac{1}{N} \sum_{j=1}^{N} \text{log} \frac{\text{exp}(\tilde{i_j} \cdot t_{N-j} / \tau)}{\sum_{k=0}^{N-1}\text{exp}(\tilde{i_j} \cdot t_{N-k} / \tau)}\right ).
\label{eq:i2t}
\end{aligned}
\end{equation}
\end{small}And the text-to-image part can be formulated as:
 \begin{small}
 \begin{equation}
 	\begin{aligned}
 	\mathcal{L}_{T2\tilde{I}} & = \lambda * \left(- \frac{1}{N} \sum_{j=1}^{N} \text{log} \frac{\text{exp}(t_j \cdot \tilde{i_j} / \tau)}{\sum_{k=1}^{N}\text{exp}(t_j \cdot \tilde{i_k} / \tau)}\right ) \\ 
 	& + (1-\lambda) * \left(- \frac{1}{N} \sum_{j=1}^{N} \text{log} \frac{\text{exp}(t_j \cdot \tilde{i}_{N-j} / \tau)}{\sum_{k=0}^{N-1}\text{exp}(t_j \cdot \tilde{i}_{N-k} / \tau)}\right ).
 	\end{aligned}
 \label{eq:t2i}
 \end{equation}
 \end{small}

\noindent \textbf{$\gamma \leq 0.5$:} We apply the mixup on the texts, and the mixed batch can be denoted as $\{(I_1, \tilde{T}_1), (I_2, \tilde{T}_2), \ldots, (I_N, \tilde{T}_N)\}$. The image-to-text matching part can be formulated as:
\begin{small}
\begin{equation}
\begin{aligned}
	\mathcal{L}_{I2\tilde{T}} & = \lambda * \left(- \frac{1}{N} \sum_{j=1}^{N} \text{log} \frac{\text{exp}(i_j \cdot \tilde{t_j} / \tau)}{\sum_{k=1}^{N}\text{exp}(i_j \cdot \tilde{t_k} / \tau)}\right ) \\ 
 	& + (1-\lambda) * \left(- \frac{1}{N} \sum_{j=1}^{N} \text{log} \frac{\text{exp}(i_j \cdot \tilde{t}_{N-j} / \tau)}{\sum_{k=0}^{N-1}\text{exp}(i_j \cdot \tilde{t}_{N-k} / \tau)}\right ).
\label{eq:i2t}
\end{aligned}
\end{equation}
\end{small}And the text-to-image part can be formulated as:
 \begin{small}
 \begin{equation}
 	\begin{aligned}
 	\mathcal{L}_{\tilde{T}2I} & = \lambda * \left(- \frac{1}{N} \sum_{j=1}^{N} \text{log} \frac{\text{exp}(\tilde{t_j} \cdot i_j / \tau)}{\sum_{k=1}^{N}\text{exp}(\tilde{t_j} \cdot i_k / \tau)}\right ) \\ 
	& + (1-\lambda) * \left(- \frac{1}{N} \sum_{j=1}^{N} \text{log} \frac{\text{exp}(\tilde{t_j} \cdot i_{N-j} / \tau)}{\sum_{k=0}^{N-1}\text{exp}(\tilde{t_j} \cdot i_{N-k} / \tau)}\right ).
 	\end{aligned}
 \label{eq:t2i}
 \end{equation}
 \end{small}Generally, the coin flipping mixup loss $\mathcal{L}_{\text{coin}}$ can be formulated as:
\begin{align}
\mathcal{L}_{\text{coin}} =
\begin{cases}
\mathcal{L}_{\tilde{I}2T} + \mathcal{L}_{T2\tilde{I}},  & \text{if } \gamma > 0.5 \\
\mathcal{L}_{I2\tilde{T}} + \mathcal{L}_{\tilde{T}2I}, & \text{if } \gamma \leq 0.5.
\end{cases}
\label{eq:mutual_lower_bound}
\end{align}


\subsection{Discussions on Sampling Strategies}
\label{subsec:sequential}

Except for random sampling and debiased sampling mentioned in the manuscript Sec.~3.1, we further explore another strategy, \ie, sequential sampling. As the name suggests, sequential sampling pre-defines the sampling order of multiple datasets and generates batches from the sequence of datasets. Illustrations of three sampling strategies are shown in Figure~\ref{fig:sampling}. 
\begin{figure}[h]
    \centering
    \includegraphics[width=0.5\linewidth]{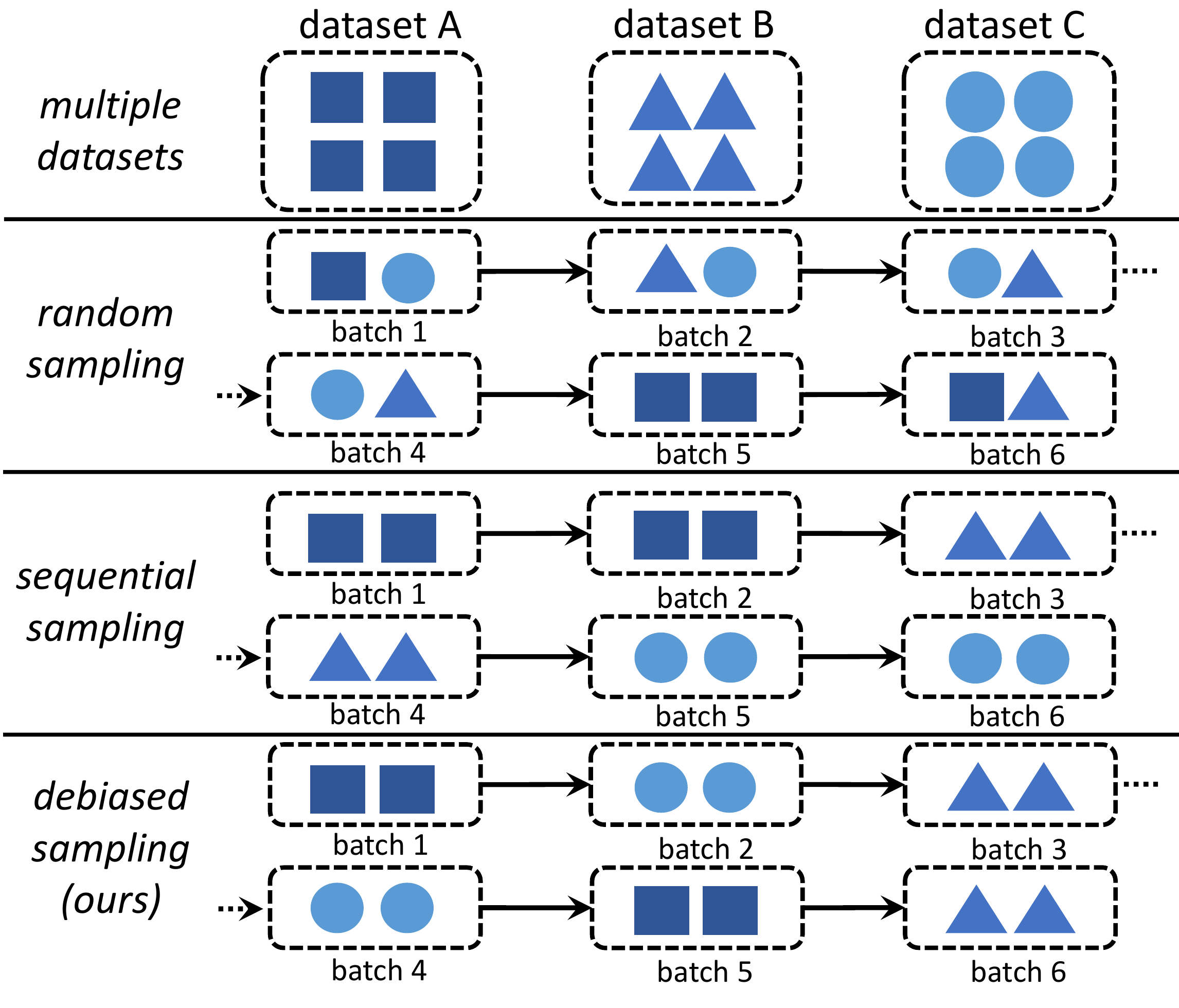}
    \caption{Comparisons between random, sequential, and debiased sampling strategies.}
    \label{fig:sampling}
\end{figure}

In the following, we study the effect of different sampling strategies. RSUM scores of zero-shot image-text retrieval task on COCO and F30K datasets are provided in Figure~\ref{fig:sampling_hist}, we notice the following phenomena on down-stream tasks:
\begin{list}{$\bullet$}{\leftmargin=1em \itemindent=0em}
	\item Sequential sampling also yields better results on downstream tasks than the random sampling.
	\item The order of datasets in sequential sampling exerts non-negligible influences on model performances. 
\end{list}
Subsequently, we further discuss these observations.

\begin{figure}[!htb]
    \centering
    \includegraphics[width=0.6\linewidth]{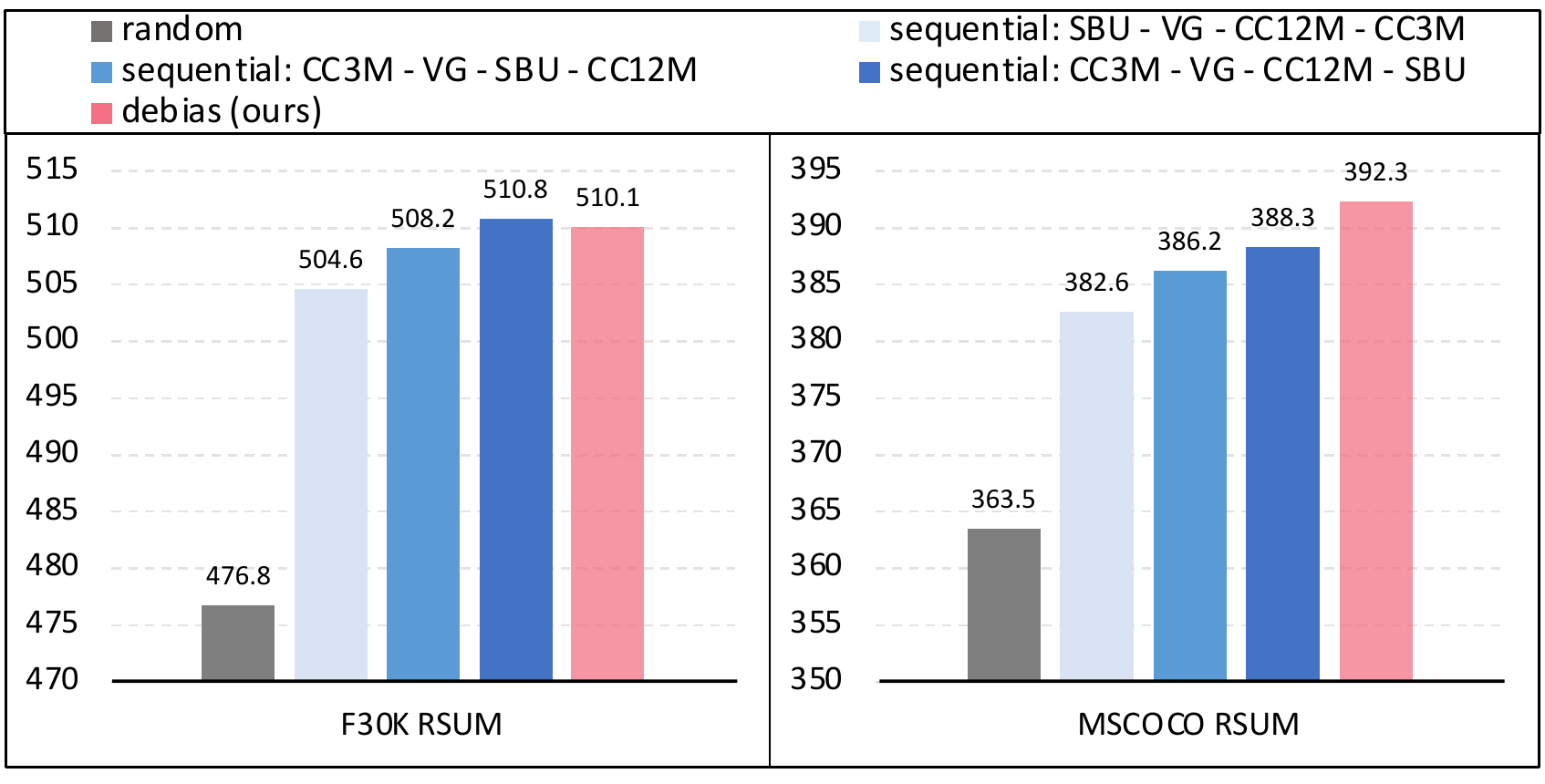}
    \caption{Comparisons between random and sequential sampling.}
    \label{fig:sampling_hist}
\end{figure}

\noindent \textit{(1)~Why sequential sampling works?}
As proven in Sec.~3 of the manuscript, debiased learning greatly benefits the contrastive vision-language pre-training. We believe that sequential sampling also tackles the dataset bias issue, since it also ensures the samples within a training batch come from one dataset. 

\noindent \textit{(2)~Why does the order matter?}
It is observed that adjusting the order of datasets in sequential sampling exerts non-negligible influences on model performances. We conjecture that the domain relevance between the ``last seen'' dataset and the downstream dataset is directly proportional to model performances, especially for zero-shot scenarios. For instance, SBU, MSCOCO, and F30K provide images with visually relevant captions, while CC3M and CC12M contain images coupled with noisy or visually irrelevant captions. Results in Figure~\ref{fig:sampling_hist} validate that setting SBU as the last dataset leads to consistently superior results, compared with CC3M or CC12M being the last one.  

\noindent \textit{(3)~Drawback of sequential sampling.} 
For sequential sampling, undoubtedly, enumerating the order of collected datasets is not acceptable for real-scenario applications, and the sequence needs further adjustment if new datasets are introduced. Our proposed debiased sampling effectively tackles this problem, and achieves better results.

\subsection{Discussions on Dataset Bias}
\label{subsec:dataset_bias}
The dataset bias could be generally divided into two types, \ie, semantic bias and context bias. Semantic bias corresponds to that semantics can be totally different between datasets, as mentioned in the manuscript. Context bias corresponds to image and text contexts, \eg, image style, caption lengths and so on. In this part, we demonstrate that debiased sampling works well on solving such context bias. 

For avoiding semantic bias, we use only CC3M in the following. We split CC3M into part A with long captions and part B with short captions. Then, we introduce image style bias by adding a red bounding box to each image of part A. We use random sampling to train a CLIP model and conduct illustrations in the Figure~\ref{fig:rebuttal}. Due to context bias, features of A~(red) and B~(blue) are separated, and gradients are inferior when training with random sampling. Debiased sampling tackles context bias and improves RSUMs from 211.8/328.0 to 239.5/367.5 on COCO/F30K.

\begin{figure}[h]
\vspace{-10pt}
    \centering
    \includegraphics[width=0.9\linewidth]{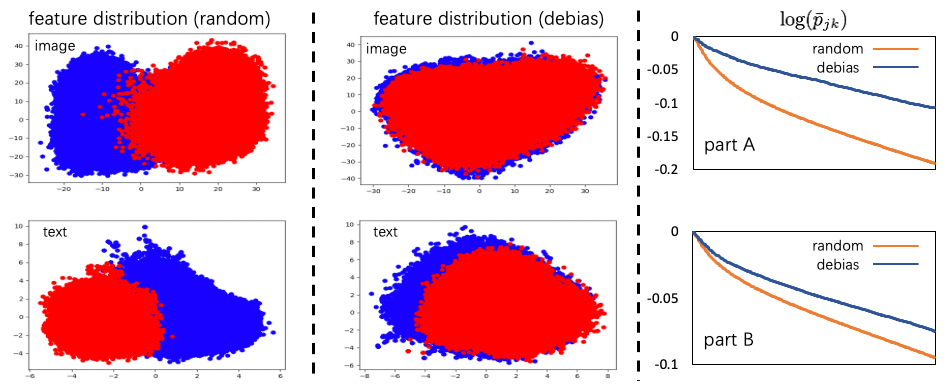}
    \caption{Effects of debiased sampling on context bias.}
    \label{fig:rebuttal}
\vspace{-20pt}
\end{figure}

Dataset bias and biased data~(feature) distribution are different concepts. Both biased feature distributions and inferior gradients are effects, while dataset bias is the cause. Dataset bias is the inherent property of training data, which exists before training. The CLIP model captures such bias, allowing the model to distinguish samples with different biases easily. Figure~3 and 4 in the manuscript both reveal bad effects of dataset bias in CLIP and prove the bias is captured by the model. 

\subsection{Application of Debiased Sampling on a Single Dataset}
\label{subsec:ds_on_one_dataset}
Debiased sampling also works well on a single source dataset. Concretely, we extract features of CC12M data, apply the KMeans on features, and produce 100 clusters. Then, we regard 100 clusters as 100 data sources, and apply debiased sampling for training a CLIP model, improving RSUMs on COCO/F30K from 370.8/502.5 to 384.4/511.1.

\subsection{More Linear Probing Results}
\label{subsec:linear_prob}
We provide linear probing results on other datasets. Six datasets are included, \ie,  CUB-200-2011~\cite{cub}~(200 categories), Food-101~\cite{food101}~(101 categories), Oxford Pets~\cite{pets}~(37 categories), FGVC Aircraft~\cite{aircraft}~(100 categories), iNaturalist-17~\cite{inat}~(5089 categories), and Places365~\cite{places}~(365 categories). Due to the pre-training dataset and linear probing hyper-parameters of CLIP is not accessible~(\eg, parameters are obtained by grid search with $\mathtt{sklearn}$). We mainly compare ZeroVL with our re-implemented CLIP. Results are reported in Table~\ref{tab:linear_prob}, and ZeroVL consistently outperforms CLIP on all datasets, which further validates the effectiveness of our baseline on linear probing tasks.

\begin{table}[h]
\centering
\setlength{\tabcolsep}{0.8mm}{
\begin{scriptsize}
\begin{tabular}{c|cc|c|c|c|c|c|c|c|c} \Xhline{2\arrayrulewidth}
\multirow{3}{*}{} &  \multicolumn{4}{c|}{pre-training} & \multicolumn{6}{c}{linear probing} \\ \cline{2-11}
 & \multicolumn{2}{c|}{computation} &  \multirow{2}{*}{data}  & input & \multirow{2}{*}{CUB} & \multirow{2}{*}{Food101} & \multirow{2}{*}{Pets} & \multirow{2}{*}{Aircraft} & \multirow{2}{*}{iNat17} & \multirow{2}{*}{Places365}  \\ 
                            &    device                 &    count   &                        & size   &   &  &    & & &     \\ \hline   
CLIP~(our impl.)                       & V100                  & 8                      & 14M  & 224 & 61.9 & 84.3 & 85.9 & 50.0 & 43.7 & 53.6                   \\
CLIP~(our impl.)                       & V100                  & 128                      & 14M  & 224 & 75.0 & 88.5 & 89.9 & 51.3   & 53.4 & 54.0                \\  
ZeroVL~(ours)                       & V100                  & 8                      & 14M  & 224 & \textbf{75.7} & \textbf{90.9} & \textbf{92.0} & \textbf{52.1} & \textbf{54.3} & \textbf{55.8}                 \\ 
\Xhline{2\arrayrulewidth}
\end{tabular}
\end{scriptsize}}
\caption{Linear probing results.}
\label{tab:linear_prob}
\end{table}

\subsection{Details of Pre-Training Datasets}
\label{subsec:datasets}
\noindent \textbf{Open-source datasets.}
 Four widely-used image-text pair datasets are selected for pre-training. Details are as followed: 
 
 \begin{itemize}
     \item SBU Captioned Photos~(SBU)~\cite{sbu} contains 1M images with associated visually relevant captions.
     \item Visual Genome~(VG)~\cite{vg} consists of around 100K images and 5M captions, where each image is coupled with 50 captions. For training efficiency, we filter 5 out of 50 captions for each image according to largest areas of bounding box regions. 
     \item Conceptual Captions 3M~(CC3M)~\cite{cc3m} contains around 3.3M images annotated with captions, collected from web data with an automatic collection pipeline. 
     \item  Conceptual 12M~(CC12M)~\cite{cc12m} is similar to CC3M and the collection pipeline is relaxed. Consequently, the data in CC12M is relatively noisier than CC3M.
 \end{itemize}
 
 A part of download links provided by CC3M and CC12M are lost. Collectively, our visual-linguistic corpus for pre-training is composed of around 14.23M image-text pairs from various domains.
 
\vspace{5pt}
\noindent \textbf{Web data.} The web data is mainly collected from an image library community Tuchong~\footnote{\href{https://www.tuchong.com}{https://www.tuchong.com}}. Due to the double-blind review policy, we are not allowed to provide the name of the community. Each image is coupled with a caption created by the image's author. The 100M web data comprises 14M academic data and 86M of web-crawling data~(from Tuchong). For applying debiased sampling, we consider the web-crawling data as a prominent source and apply debiased sampling on datasets from five sources.



\subsection{Training Details}
\label{subsec:training_detail}

We elaborate the training details of our strong baseline.

\vspace{5pt}
\noindent \textbf{Data preparation.} 
Batches are comprised by applying the debaised sampling strategy on academic pre-training datasets, \ie, SBU, VG, CC3M and CC12M. Each image is randomly cropped to a rectangular region with aspect ratio sampled in $[3/4,4/3]$ and area sampled in $[60\%,100\%]$, then resized to 224$\times$224 resolution. Regarding the corresponding text, we set the max length to 25 and use a percentage of 20\% input words for processing. For each word, we mask it, replace it with a random word, or delete it with a probability of 50\%, 10\% and 40\%, respectively. We directly apply AutoAugment after crop operation and the policy is search on ImageNet~\footnote{\href{https://github.com/4uiiurz1/pytorch-auto-augment}{https://github.com/4uiiurz1/pytorch-auto-augment}}. Coin flipping mixup is also used in the training phase, and the $\alpha$ is set to 0.1 in the coin flipping mixup. During test, images are resized to 256$\times$256 and center cropped to 224$\times$224, while no specific process is applied to texts. 

\vspace{5pt}
\noindent \textbf{Model architecture.} Image and text encoders are ViT-B/16 and BERT-Base, respectively. The image encoder is pre-trained on ImageNet~\cite{imagenet} which could be directly obtained from the $\texttt{timm}$~\footnote{\href{https://github.com/rwightman/pytorch-image-models}{https://github.com/rwightman/pytorch-image-models}} library while the text encoder is pre-trained on BookCorpus~\cite{bookcorpus} and English Wikipedia from the $\texttt{HuggingFace}$~\footnote{\href{https://huggingface.co}{https://huggingface.co}} library. \texttt{[CLS]} tokens from image and text encoders are extracted and then projected to 512-dim compact embeddings and $\ell$-2 normalized for calculating the contrastive loss.

\vspace{5pt}
\noindent \textbf{Training.}
AdamW optimizer is used for training and the weight decay is 1e-3. Based on our decoupled gradient accumulation, the dual-encoder model is trained for 20 epochs on 8 Nvidia V100 GPUs with a batch size of 16,384. The learning rate is initialized to 1e-4 and follows a cosine decay schedule. Notably, we set a minimum learning rate 1e-5 to avoid over-fitting. The embedding dimension for image and text representations is 512 and the trainable temperature of contrastive loss is initialized to 0.02.

\end{alphasection}

\clearpage
%
%
\bibliographystyle{splncs04}
\bibliography{egbib}
\end{document}